\documentclass[10pt,twocolumn,letterpaper]{article}
\usepackage[pagenumbers]{cvpr} 
\usepackage{ulem}
\usepackage{graphicx}
\usepackage{amsmath}
\usepackage{amssymb}
\usepackage{booktabs}
\usepackage[ruled, lined, linesnumbered, longend]{algorithm2e}
\usepackage{multirow}
\usepackage{enumitem}
\usepackage[numbers,sort&compress]{natbib}
\usepackage[pagebackref,breaklinks,colorlinks]{hyperref}
\newcommand{\nickname}{U-Next}

\usepackage[capitalize]{cleveref}
\crefname{section}{Sec.}{Secs.}
\Crefname{section}{Section}{Sections}
\Crefname{table}{Table}{Tables}
\crefname{table}{Tab.}{Tabs.}

\begin{document}

\title{Small but Mighty: Enhancing 3D Point Clouds Semantic Segmentation \\ with U-Next Framework}

\author{Ziyin Zeng$^{1}$, Qingyong Hu$^{2}$, Zhong Xie$^{1}$, Jian Zhou$^{3}$ and Yongyang Xu$^{1}\thanks{Corresponding author}$ \\
$^1$China University of Geosciences, $^2$University of Oxford, $^3$Wuhan University\\
{\tt\small \{zengziyin, xiehzhong, yongyangxu\}@cug.edu.cn, qingyong.hu@cs.ox.ac.uk,  jianzhou@whu.edu.cn}}
\maketitle

\begin{abstract}

We study the problem of semantic segmentation of large-scale 3D point clouds. In recent years, significant research efforts have been directed toward local feature aggregation, improved loss functions and  sampling strategies. While the fundamental framework of point cloud semantic segmentation has been largely overlooked, with most existing approaches rely on the U-Net architecture by default. In this paper, we propose U-Next, a small but mighty framework designed for point cloud semantic segmentation. The key to this framework is to learn multi-scale hierarchical representations from semantically similar feature maps. Specifically, we build our U-Next by stacking multiple U-Net $L^1$ codecs in a nested and densely arranged manner to minimize the semantic gap, while simultaneously fusing the feature maps across scales to effectively recover the fine-grained details. We also devised a multi-level deep supervision mechanism to further smooth gradient propagation and facilitate network optimization. Extensive experiments conducted on three large-scale benchmarks including S3DIS, Toronto3D, and SensatUrban demonstrate the superiority and the effectiveness of the proposed U-Next architecture. Our U-Next architecture shows consistent and visible performance improvements across different tasks and baseline models, indicating its great potential to serve as a general framework for future research.

\end{abstract}

\section{Introduction}

Thanks to the substantial development of 3D acquisition equipment, 3D point cloud analysis has attracted increasing attention in recent years \cite{guo2020deep}. As a fundamental step of 3D scene understanding, point cloud semantic segmentation is aimed at assigning a unique semantic label to each point based on its geometrical structure. This task has garnered significant research interest owing to its wide applications in autonomous driving, robotics, and virtual reality \cite{SemanticKITTI, pcl, blanc2020genuage}. However, it remains highly challenging to achieve fine-grained semantic scene parsing since the acquired point clouds are naturally orderless, irregular, and unstructured, coupled with large-scale points, varying-size objects, and non-uniform distributions.

\begin{figure}[!t]
\vspace{-1.0em}
\centering
\includegraphics[width=0.47\textwidth]{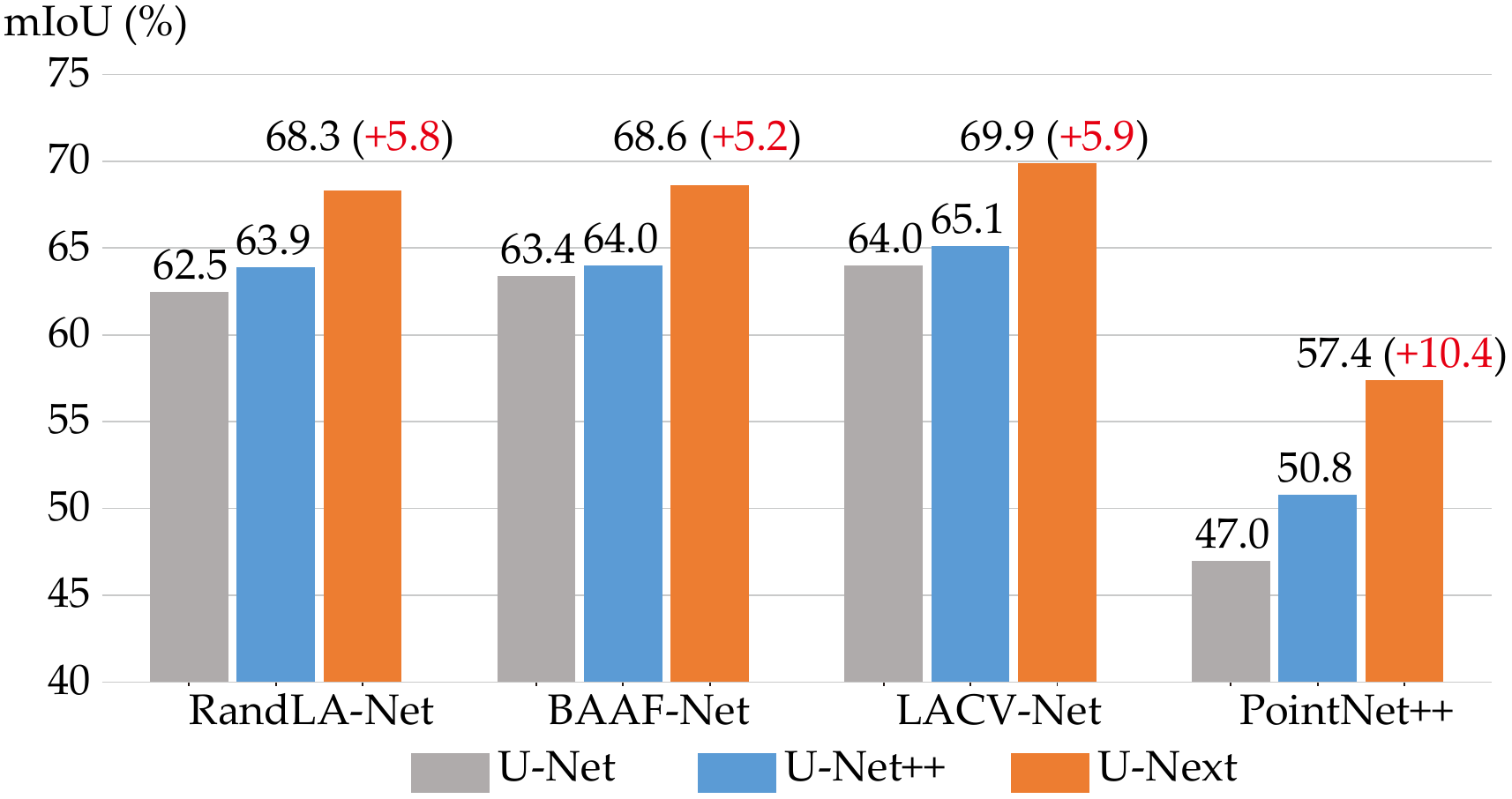}
\caption{Performance of various algorithms on S3DIS dataset (Area 5) with different frameworks. Our U-Next framework demonstrated a notable advantage over both U-Net and U-Net++.}
\label{fig1}
\vspace{-1.0em}
\end{figure}

Recently, the pioneering research work PointNet \cite{PointNet} has been proposed to directly learn from unstructured 3D point clouds, while with high efficiency and encouraging performance on several downstream tasks. Later, deep learning on point cloud analysis \cite{guo2020deep} has been extensively explored. A plethora of sophisticated neural architectures \cite{PointNet++, PointASNL, hu2021learning, ao2022you, BAAF-Net, hu2022sqn, GACNet, KPpyramid, KPConv, SparseConv} have been proposed, leading to consistent improvements in performance on large-scale 3D point cloud benchmarks \cite{s3dis, Semantic3D, SemanticKITTI, hu2022sensaturban, chen2022stpls3d}. Despite the promising progress achieved, existing approaches have primarily focused on representation learning, \textit{i.e.,} learning an effective representation for downstream tasks of 3D point clouds, including point-based \cite{PointNet, PointNet++, RandLA-Net, PointNext, PointWeb}, projection-based \cite{boulch2017unstructured, kundu2020virtual}, voxel-based \cite{tchapmi2017segcloud, SparseConv, 4dMinkpwski}, and point-voxel-based techniques \cite{e3d, Point-Voxel_CNN}. Furthermore, several methods have extensively explored various local feature aggregation mechanisms \cite{RandLA-Net, PointASNL, DeepGCNs} and efficient neural network architectures \cite{SparseConv, 4dMinkpwski}. However, few endeavors were made to improve the segmentation framework in the field of 3D point cloud learning. Most existing approaches tend to indiscriminately employ the widely-used U-Net architecture, while overlooking its inherent limitations, which have been extensively documented in the context of 2D natural and medical images.

As one of the most widely used neural architectures, U-Net \cite{UNet} is built upon the encoder-decoder structure, and the key is to use skip connections to combine intermediate latent feature maps from encoder and decoder. Although this architecture appears symmetrical and natural, the combined same-scale feature maps are actually far from semantically similar (\textit{shallow} and \textit{low-level} features from encoder +  \textit{deep} and \textit{semantic} features from decoder), and there is no solid theory to prove such a combination is optimal \cite{UNet++}. On the other hand, the optimal depth of the network is also apriori unknown, which requires manual architecture tuning. To this end, a handful of approaches \cite{ResUNet,DenseUNet,TransUNet,UCTransNet,ResUNet,DenseUNet} have started to explicitly improve the U-Net framework, including U-Net++ \cite{UNet++} with nested, dense re-designed skip pathways, UNet3+ \cite{UNet3} with full-scale skip connections, and TransUNet \cite{TransUNet} with transformer as encoder, \textit{etc}. Although proven to be highly effective in the segmentation of medical images, it remains an open question whether these improved architectures are \textit{equally effective} in 3D point cloud segmentation, especially considering the internal differences between 3D point clouds and images.

To answer this question, we first conduct an in-depth analysis of U-Net and its variants, followed by exploratory experiments to verify whether existing architectures can be smoothly tailored to 3D point clouds. Unfortunately, the U-Net++ architecture, which has been successful in 2D image segmentation, did not yield significant improvements in the segmentation of 3D point clouds (as shown in Figure \ref{fig1}). This can be attributed to the orderless and irregular nature of 3D point clouds, whereby significant information loss occurs  during aggressive upsampling and downsampling, especially considering most existing approaches usually adopt primitive downsampling (random sampling or farthest point sampling), and upsampling scheme (nearest interpolation, and trilinear interpolation). Moreover, within the overarching framework of the network architecture, the successive reduction (high-to-low) and enlargement (low-to-high) of feature dimensions can induce significant noise that accumulates over layers and has a profound impact on segmentation accuracy. In this case, simply aggregating features at varying scales may not be conducive to segmentation accuracy, but increasing the difficulty of optimization due to the sizeable differences in features between different levels.

Bearing this in mind, we hereby propose a general, conceptually-simple, yet highly effective point cloud semantic segmentation architecture in this paper, termed \nickname{}. The key of our \nickname{} is to leverage the most basic U-Net $L^1$ sub-network, as shown in Figure \ref{fig:unext}-(B), which only performs the downsampling and upsampling operations once, thus has the minimal semantic gap and no extra noise, as the building blocks to build the \nickname{} architecture. That is, stacking as many U-Net $L^1$ codecs as possible to learn point features from different scales, and enabling the local feature maps free to flow laterally, upward, or downward, as shown in Figure \ref{fig:unext}. The incorporation of additional scales in our framework serves to strengthen the representation of the input data, thereby resulting in an improved overall accuracy. Finally, we apply multi-level deep supervision at each intermediate node, easing the difficulty in learning and optimization of each constituent U-Net $L^1$ sub-network. Experiments on three large-scale public benchmarks, including S3DIS \cite{s3dis}, Toronto3D \cite{Toronto3D}, and SensatUrban \cite{SensatUrban}, show that our framework can lead to 2.2\% $\sim$ 10.1\% improvements in mIoU score, based on the widely-used RandLA-Net network. Moreover, our framework can be seamlessly integrated into existing segmentation networks, such as PointNet++ \cite{PointNet++}, BAAF-Net \cite{BAAF-Net}, and LACV-Net \cite{LACV-Net}, resulting in improved segmentation accuracy. Remarkably, these improvements were achieved without incurring any visible additional computational costs, thus highlighting the versatility and effectiveness of our U-Next framework. Overall, our key contributions can be summarized as:

\begin{itemize}[leftmargin=*]
\setlength{\itemsep}{0pt}
\setlength{\parsep}{0pt}
\setlength{\parskip}{0pt}
\item We provide an in-depth analysis of U-Net architecture evolution in the field of 3D point clouds, identifying U-Net $L^1$ sub-network as a suitable component for fine-grained point cloud segmentation.
\item We propose a general and effective segmentation architecture \nickname{}, by stacking multiple fundamental U-Net $L^1$ codecs with multi-level deep supervision.
\item Extensive experiments on three large-scale benchmarks and up to four baseline approaches demonstrate the effectiveness and generality of our \nickname{} architecture.
\end{itemize}

\section{Related Work}

\subsection{Learning to Segment 3D Point Clouds}
To analyze unstructured 3D point clouds, early works usually utilize hand-crafted feature descriptors \cite{tradition_1,tradition_2}, while recent works are mainly based on data-driven deep neural architectures. Here, we briefly review the related learning-based techniques, including projection-based, voxel-based, and point-based methods.

Considering the orderless and unstructured nature of 3D point clouds, directly applying standard convolution neural networks on 3D point clouds remains challenging. Consequently, projection-based \cite{PB_1,PB_2,PB_3} and voxel-based methods \cite{VB_1,VB_2,VB_3} are proposed to first convert unstructured point clouds into intermediate regular 2D grid images or 3D dense voxels through projection or voxelization steps, and then leverage the successful CNNs to learn from the structured representation. Albeit promising results, the main drawback of these techniques lies in the information loss caused by 3D-2D projection or voxelization steps, especially for learning fine-grained features for large-scale 3D point clouds. By contrast, point-based methods are designed to directly operate on irregular point clouds without using intermediate representations. The pioneering PointNet \cite{PointNet} is proposed to learn point-wise feature representations by leveraging the shared MLPs. Later, an increasing number of point-based methods have been proposed to learn from unstructured point clouds \cite{PointNet++,AD-SAGC,KPConv,RandLA-Net,BAAF-Net,PointWeb,PointASNL,SCF-Net,BAF-LAC,LEARD}. Several representative research works have incorporated Recurrent Neural Networks (RNNs) \cite{3P-RNN,RSNet}, Graph Neural Networks (GNNs) \cite{SPG,GACNet,DeepGCNs}, and Transformer architectures \cite{PCT,PT,PointASNL}, to achieve better performance on different tasks.

\begin{figure}[t]
\centering
\includegraphics[width=0.47\textwidth]{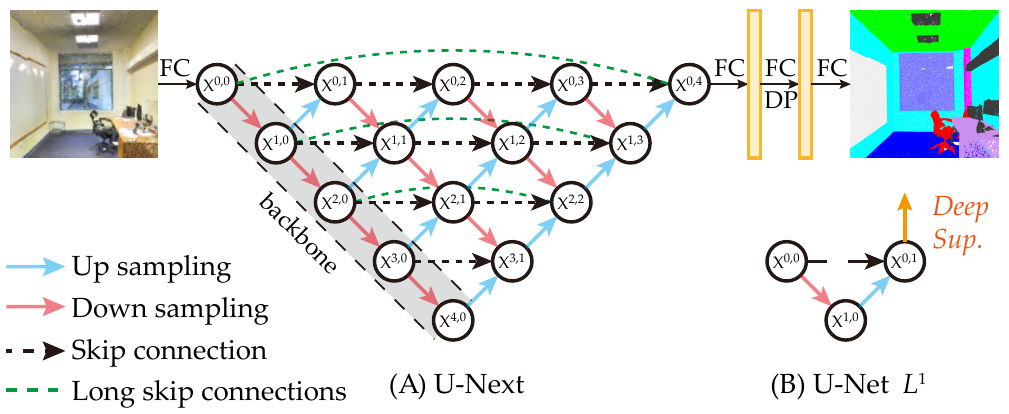}
\caption{Illustration of the proposed U-Next and U-Net $L^1$. (A) The detailed architecture of the proposed U-Next. (B) The U-Net $L^1$ sub-network with deep supervision. FC: Fully Connected layer; DP: Dropout; Deep Sup.: Deep Supervision.}
\label{fig:unext}
\vspace{-1.0em}
\end{figure}

\subsection{Multi-Scale Feature Fusion}
Multi-scale fusion has been widely studied and successfully applied in 2D image processing \cite{MF_1,MF_2,MF_3,MF_4,HRNet}. Existing works are mainly built upon the encoder-decoder framework, with different-scale or different-resolution feature maps in the encoder or decoder sub-networks. Summation or concatenation operations are usually used to fuse different feature maps \cite{MF_a_1,MF_a_2,MF_a_3,MF_a_4}. To effectively analyze large-scale 3D point clouds which are composed of varying-size objects, a handful of research works also started to explore multi-scale feature fusion in 3D point clouds. The pioneering PointNet++ \cite{PointNet++} explores multi-scale grouping and abstraction to aggregate features from different scales. The adaptive fusion module proposed by BAAF-Net \cite{BAAF-Net} is used to fuse fine-grained representations from the multi-resolution feature maps adaptively.

\subsection{Learning with Deep Supervision}
Deep supervision was first proposed to tackle the issues of gradient vanishing and slow convergence speed of training deep neural networks \cite{Lee,GoogLeNet}. As an effective training technique, it has been also introduced to different tasks to improve performance \cite{DS_1,DS_2,DS_3,HRNet}. Lee et al. \cite{Lee} demonstrated that deeply supervised layers can improve the learning capacity of the hidden layer, further enforcing the intermediate layers to learn discriminative features, and enabling fast convergence and regularization of the network. Dou et al. \cite{Dou} introduce a deep supervision paradigm to deal with the optimization issues during network training by combining predictions from varying resolutions of feature maps. Deep supervision can also be used in a U-Net-like architecture, where the intermediate feature maps are treated as the output of sub-networks. U-Net++ \cite{UNet++} designs a deep supervision scheme with a redesigned skip connection in U-Net architecture to further improve the quality of feature maps. U-Net3+ \cite{UNet3} achieves deep supervision by learning from aggregated feature maps hierarchically.

\subsection{Improving U-Net Architecture}
The original U-Net \cite{UNet}, which is proposed with symmetrical encoder-decoder architecture, is widely used in image segmentation. In particular, skip connections are used to combine the high-level semantic features from the decoder and low-level details from the encoder. Based on this architecture, a number of improved network architectures have been further proposed. Res-UNet \cite{ResUNet} and Dense-UNet \cite{DenseUNet} replace each sub-module of the U-Net with a residual connection or a dense connection, respectively. U-Net++ \cite{UNet++} redesigns skip connection pathways to explore the optimal depth by fusing multiple U-Net sub-networks with varying depths, where each sub-network is connected through a series of nested, dense skip pathways. U-Net3+ \cite{UNet3} proposes full-scale skip connections to fully exploit multi-scale feature representations. TransUNet \cite{TransUNet} introduces Transformer into U-Net to capture the best of both models. This method not only encodes strong global context, but leverages low-level CNN features.

Similarly, this paper also aims to fundamentally improve the U-Net architecture for 3D point cloud segmentation. However, different from  existing techniques, we first systematically analyze the inherent drawbacks of existing architectures, and then propose our U-Next architecture by stacking a maximum number of sub-networks with minimal codec semantic gap, to effectively learn from non-uniform and unstructured 3D point clouds.
\begin{figure*}[!t]
\vspace{-1.5em}
\centering
\includegraphics[width=0.9\textwidth]{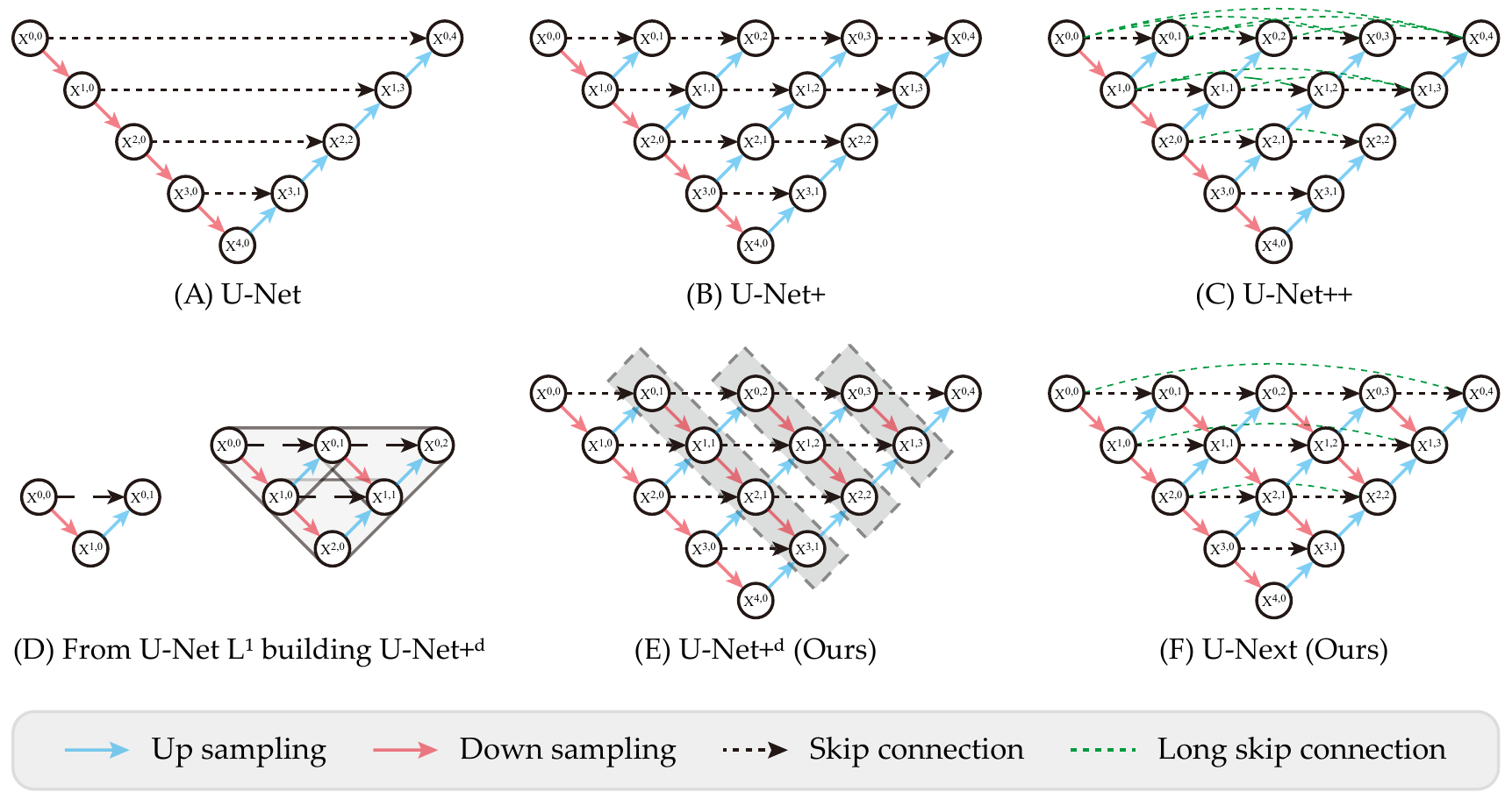}
\caption{Illustration of the evolution from U-Net to U-Next. The key contributions of each architecture are identified. (A) is the original U-Net, (B) and (C) are U-Net+ and U-Net++ proposed by \cite{UNet++}, (D) shows how to build U-Net+$^d$ from U-Net $L^1$, and (E) and (F) are the proposed U-Net+$^d$ and U-Next. U-Net+ has more up-sampling branches than the original U-Net for finer granularity. U-Net++ uses dense long skip pathways based on U-Net+ to aggregate features of the same resolution. U-Net+$^d$ has more down-sampling branches and repeatedly fuses multi-scale feature maps with sub-networks to recover fine-grained details. U-Next employs a long skip connection in each horizontal level based on U-Net+$^d$, focusing on elementary features.}
\label{fig:evo}
\vspace{-1.0em}
\end{figure*}

\section{The Proposed \nickname{}}
\label{sec:metho}

\subsection{Revisit U-Net-Like Architectures}

U-Net-like architectures have been widely used in modern segmentation models. As shown in Figure~\ref{fig:evo}(A), this network is based on a symmetrical encoder-decoder architecture, with skip connections playing a crucial role in integrating the deep, semantically-rich feature maps from the decoder sub-network with the shallow, fine-grained feature maps from the encoder sub-network. Despite achieving promising results in the recovery of fine-grained details for dense prediction tasks like semantic and instance segmentation, this architecture has limitations. Specifically, it only performs feature aggregation at the same scale, which is restrictive and lacks a solid theoretical foundation. Furthermore, the combined feature maps have been shown to be semantically dissimilar \cite{UNet++}.

\noindent \textbf{Challenges.} Recently, a handful of advanced architectures such as U-Net+ (Figure~\ref{fig:evo}(B)) and U-Net++ \cite{UNet++} (Figure~\ref{fig:evo}(C)) have been proposed. These architectures assemble various levels of U-Net as sub-networks with deep supervision to effectively leverage multi-scale feature maps, and make efforts in reducing semantic gaps between combined feature maps in medical image segmentation. Nevertheless, the application of these architectures to 3D point clouds remains a highly challenging task. \textbf{First}, most existing architectures can be fundamentally viewed as an ensemble of U-Nets with varying different depths. That is, the semantic gap between the combined features in deeper U-Nets remains substantial, potentially hindering network learning and optimization. \textbf{Second}, Considering aggressive upsampling and downsampling are frequently used in point cloud neural architectures, coupled with the orderless and unstructured nature of 3D point clouds, the recovered full-resolution feature maps from horizontal and low-to-high features fusion may still be inadequate for fine-grained semantic segmentation.

\noindent \textbf{Solutions.} 
Given the inevitable information loss during downsampling and upsampling, along with the growing semantic gap between codecs with increasing U-Net depth, we were motivated to re-examine the U-Net architecture and propose a dedicated framework for point cloud semantic segmentation. Drawing inspiration from the concept of building blocks, we wondered if it was feasible to build U-Net architecture from the fundamental U-Net $L^1$ sub-network (Figure~\ref{fig:evo}(D)), since there is a single downsampling and upsampling operation in U-Net $L^1$ sub-network, resulting in a naturally small semantic gap. Next, considering the shallow structure and limited learning capacity of the U-Net $L^1$ sub-network, we adopt a straightforward approach to enhance the feature learning capacity by stacking multiple U-Net $L^1$ codecs in a nested and densely arranged manner. This allows the network to learn multi-scale features, ultimately leading to improved performance. Additionally, to ensure smooth gradient backpropagation and reduce optimization difficulty during training, it is desirable to further introduce a dedicated deep supervision scheme. To summarize, a feasible solution is to take the fundamental U-Net L1 codecs as building blocks for our framework, which can simultaneously minimize the semantic gap and recover fine-grained feature maps, thus enabling high-quality 3D semantic segmentation.

\subsection{U-Next: Building Backbone Architecture with U-Net L$^1$ Sub-Networks}

Before building the final U-Next framework, we first proposed a hierarchically ensemble architecture based on the fundamental U-Net $L^1$ codecs, as shown in Figure ~\ref{fig:evo}(E). It can be seen that this architecture explicitly adds the connections between different-level decoder nodes, compared with the previous U-Net+ architecture, enabling each node receives information from other parallel and adjacent (lower and higher) nodes, leading to sufficient and comprehensive information fusion. This architecture can be also viewed as progressively stacking as many U-Net $L^1$ codec as possible, which naturally has the minimal codec semantic gap. Further, we enriched U-Net+$^d$ architecture with long skip connections, to build the final U-Next architecture, as shown in Figure~\ref{fig:evo}(F). Different from the nested, dense skip pathways used in U-Net++ framework, our \nickname{} architecture only adds simple sparse skip connections similar to the original U-Net architecture. The detailed description of the networks are summarized as follows.

\vspace{1.5mm}
\noindent \textbf{Building U-Net+$^d$ from U-Net $L^1$:} As shown in Figure~\ref{fig:evo}(E), each node in U-Net+$^d$ receives information from other parallel and adjacent (lower and higher) nodes, leading to comprehensive and sufficient multi-scale information fusion. Specifically, We use $x^{i, j}$ to denote the output of node $X^{i, j}$ (each node in the same horizontal row with the same $i$, and in the same slash along the down-sampling direction with the same $j$). The stack of feature maps in U-Net+$^d$ represented by $x^{i, j}$ can be formulated as follows:

\begin{footnotesize}
\label{eq:UNet+d}
\begin{equation}
\setlength{\abovedisplayskip}{5pt}
x^{i,j}=
\begin{cases}
\mathcal{C}[x^{i-1,j}], & j=0\\
\mathcal{C}[x^{i-1,j},\mathcal{U}(x^{i-1,j+1})], & j>0,i=0\\
\mathcal{C}[x^{i-1,j},\mathcal{U}(x^{i-1,\ j+1}),\mathcal{D}(x^{i-1,j-1})], & j>0,i>0
\end{cases}
\setlength{\belowdisplayskip}{13pt}
\end{equation}
\end{footnotesize}

\noindent where $\mathcal{C}$(·) is coding block (usually a feature extraction module), $\mathcal{D}$(·) and $\mathcal{U}$(·) denote a down-sampling operation and an up-sampling operation respectively, and [·] denotes the concatenation operation. 

\vspace{1.5mm}
\noindent \textbf{Building U-Next from U-Net+$^d$:} We build U-Next architecture based on U-Net+$^d$ by introducing long skip connections. As shown in Figure~\ref{fig:evo}(F), the earlier feature maps are travel through the long skip connections of U-Next to integrate with the later feature maps. Specifically, concatenating feature maps with additional long skip connections are computed by fusing the first node in $i$-th row with the node $X^{i,j}$, if and only if node $X^{i,j}$ is the last node in row $i$, as follows:

\begin{footnotesize}
\label{eq:UNext}
\begin{equation}
\setlength{\abovedisplayskip}{5pt}
x^{i,j}=
\begin{cases}
\mathcal{C}[x^{i-1,j},\mathcal{U}(x^{i-1,j+1}),x^{i,0}], & i=0\\
\mathcal{C}[x^{i-1,j},\mathcal{U}(x^{i-1,\ j+1}),\mathcal{D}(x^{i-1,j-1}),x^{i,0}], & i>0
\end{cases}
\setlength{\belowdisplayskip}{13pt}
\end{equation}
\end{footnotesize}

\subsection{Learning with Multi-Level Deep Supervision}

\begin{figure}[]
\centering
\includegraphics[width=0.45\textwidth]{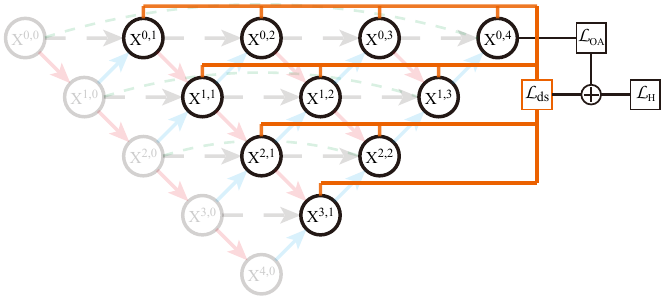}
\caption{Illustration of deep supervision. All the sub-networks are supervised. $\mathcal{L}_{ds}$: loss of the multi-level deep supervision; $\mathcal{L}_{OA}$: loss of the overall network; $\mathcal{L}_{H}$: hybrid loss.}
\label{fig:DS}
\vspace{-1.0em}
\end{figure}

Up to here, multiple U-Net $L^1$ sub-networks have been hierarchically stacked in our framework. To further learn hierarchical representations from the aggregated feature maps, we introduce multi-level deep supervision into our framework from the perspective of optimization and learning. As shown in  Figure~\ref{fig:DS}, different from U-Net++ only apply deep supervision on the generated full-resolution feature maps, our U-Next framework have explicitly applied deep supervision to each decoder node at different levels (\textit{i.e.,} for each U-Net $L^1$ sub-network). By introducing ground-truth supervision signal at different decoding stages of our framework, smooth gradient propagation and easy optimization of the network are more likely to be achieved. Therefore, all the constituent U-Net $L^1$ sub-network are trained simultaneously to improve the overall learning capacity of multi-scale feature maps. 

\vspace{1.5mm}
\noindent \textbf{Loss of multi-level deep supervision: } To achieve multi-level deep supervision, the output of decoder in \textit{each} U-Net $L^1$ sub-network is fed into a plain $1 \times 1$ convolution layer to align with the ground truth labels. After that, we follow the widely-used \textit{cross-entropy loss} to calculate the loss of each U-Net $L^1$ sub-network. Finally, the total loss of deep supervision is defined as the average of all the losses of the U-Net $L^1$ sub-networks, as follows: 

\begin{equation}
\setlength{\abovedisplayskip}{0pt}
\mathcal{L}_{ds}(Y,P)=-\frac{1}{N}\sum_{i=0}^{L} \sum_{j=0}^{L-i} \sum_{c=0}^{C} y_c^i\log{p_c^{i,j}}
\setlength{\belowdisplayskip}{0pt}
\end{equation}

\noindent where $y_c^i$ is the true label vector of the $i$-th row, $p_c^{i,j}$ is the predicted label vector of node $X^{i,j}$, $C$ is the number of label categories, $L$ is the number of architecture levels, and $N$ is the number of decoder nodes. In the multi-level deep supervision, all the sub-networks are trained, enhancing the learning ability of each sub-network for local features.

\begin{table*}[]
\centering
\setlength{\abovecaptionskip}{0cm}
\setlength{\belowcaptionskip}{-0.2cm}
\caption{Quantitative results of different approaches on the S3DIS dataset \cite{s3dis} (six-fold cross-validation).}
\resizebox{0.95\textwidth}{!}{%
\begin{tabular}{rccccccccccccccc}
\bottomrule
\multirow{2}*{Method}& \multirow{2}*{OA}& \multirow{2}*{mIoU}& \multirow{2}*{ceil.}& \multirow{2}*{floor}& \multirow{2}*{wall}& \multirow{2}*{beam}& \multirow{2}*{col.}& \multirow{2}*{wind.}& \multirow{2}*{door}& \multirow{2}*{table}& \multirow{2}*{chair}& \multirow{2}*{sofa}& \multirow{2}*{book.}& \multirow{2}*{board}& \multirow{2}*{clut.}\\
&&&&&&&&&&&&&&\\
\hline
PointNet \cite{PointNet} & 78.6 & 47.6 & 88.0 & 88.7 & 69.3 & 42.4 & 23.1 & 47.5 & 51.6 & 54.1 & 42.0 & 9.6 & 38.2 & 29.4 & 35.2 \\
DGCNN \cite{DGCNN} & 84.5 & 55.5 & 93.2 & 95.9 & 72.8 & 54.6 & 32.2 & 56.2 & 50.7 & 62.8 & 63.4 & 22.7 & 38.2 & 32.5 & 46.8 \\
SPGraph \cite{SPG} & 86.4 & 62.1 & 89.9 & 95.1 & 76.4 & 62.8 & 47.1 & 55.3 & 68.4 & 73.5 & 69.2 & 63.2 & 45.9 & 8.7 & 52.9 \\
DeepGCNs \cite{DeepGCNs} & 85.9 & 60.0 & 93.1 & 95.3 & 78.2 & 33.9 & 37.4 & 56.1 & 68.2 & 64.9 & 61.0 & 34.6 & 51.5 & 51.1 & 54.4 \\
PointWeb \cite{PointWeb} & 87.3 & 66.7 & 93.5 & 94.2 & 80.8 & 52.4 & 41.3 & 64.9 & 68.1 & 71.4 & 67.1 & 50.3 & 62.7 & 62.2 & 58.5 \\
ShellNet \cite{ShellNet} & 87.1 & 66.8 & 90.2 & 93.6 & 79.9 & 60.4 & 44.1 & 64.9 & 52.9 & 71.6 & \textbf{84.7} & 53.8 & 64.6 & 48.6 & 59.4 \\
KPConv \cite{KPConv} & -  & 70.6 & 93.6 & 92.4 & 83.1 & 63.9 & 54.3 & 66.1 & 76.6 & 57.8 & 64.0 & 69.3 & 74.9 & 61.3 & 60.3 \\
BAAF-Net \cite{BAAF-Net} & 88.9 & 72.2 & 93.3 & 96.8 & 81.6 & 61.9 & 49.5 & 65.4 & 73.3 & 72.0 & 83.7 & 67.5 & 64.3 & 67.0 & 62.4 \\
BAF-LAC \cite{BAF-LAC} & 88.2 & 71.7 & 92.5 & 95.9 & 81.3 & 63.2 & 57.8 & 63.0 & \textbf{79.9} & 70.3 & 74.6 & 60.6 & 67.2 & 65.3 & 60.4 \\
SCF-Net \cite{SCF-Net} & 88.4 & 71.6 & 93.3 & 96.4 & 80.9 & 64.9 & 47.4 & 64.5 & 70.1 & 71.4 & \textbf{81.6} & 67.2 & 64.4 & 67.5 & 60.9 \\
LACV-Net \cite{LACV-Net} & 89.7 & 72.7 & 94.5 & 96.7 & 82.1 & 65.2 & 48.6 & 69.3 & 71.2 & 72.7 & 78.1 & 67.3 & 67.2 & 70.9 & 61.6 \\
CBL \cite{CBL} & 89.6 & 73.1 & 94.1 & 94.2 & \textbf{85.5} & 50.4 & \textbf{58.8} & \textbf{70.3} & 78.3 & \textbf{75.7} & 75.0 & \textbf{71.8} & \textbf{74.0} & 60.0 & 62.4 \\
Point Transformer \cite{PT} & 90.2 & 73.5 & 94.3 & \textbf{97.5} & 84.7 & 55.6 & 58.1 & 66.1 & 78.2 & 77.6 & 74.1 & 67.3 & 71.2 & 65.7 & \textbf{64.8}\\
PointNeXt \cite{PointNext} & \textbf{90.3} & \textbf{74.9} &-&-&-&-&-&-&-&-&-&-&-&-&- \\
\hline
RandLA-Net \cite{RandLA-Net} & 88.0 & 70.0 & 93.1 & 96.1 & 80.6 & 62.4 & 48.0 & 64.4 & 69.4 & 69.4 & 76.4 & 60.0 & 64.2 & 65.9 & 60.1 \\
\textbf{\textcolor{red}{+ U-Next}} & 89.5 ${\textcolor{red}{\textbf{+1.5}}}$ & 73.2 ${\textcolor{red}{\textbf{+3.2}}}$ & 93.6 & 96.9 & 84.2 & \textbf{66.1} & 54.6 & 67.6 & 75.5 & 73.6 & 74.5 & 62.9 & 66.2 & \textbf{74.0} & 61.7 \\
\bottomrule
\end{tabular}%
\label{tab:s3dis}
}
\vspace{-0.5em}
\end{table*}

\vspace{1.5mm}
\noindent \textbf{Hybrid loss: } As shown in Figure~\ref{fig:DS}, after the output of the final node ($X^{0,4}$), the fully-connected layers are used to predict semantic labels of all input points. Therefore, we also use cross-entropy loss to calculate the loss of the overall network by the final prediction results. Further, the final loss function of the framework is a hybrid loss which combined the loss of the multi-level deep supervision and overall network, as follows: 

\begin{equation}
\mathcal{L}_{H}=\mathcal{L}_{ds} + \mathcal{L}_{oa}
\end{equation}

\section{Experiments}

\subsection{Experiment Setup}
To comprehensively evaluate the effectiveness of the proposed U-Next, we conduct experiments on the widely-used indoor S3DIS dataset \cite{s3dis}, outdoor LiDAR-based Toronto3D dataset \cite{Toronto3D}, and urban-scale photogrammetric SensatUrban dataset \cite{SensatUrban}. To quantitatively evaluate the segmentation performance, we take the per-class Intersection over Union (IoUs), mean IoU (mIoU), and Overall Accuracy (OA) as evaluation metrics. We also provide the performance of our U-Next framework in the instance segmentation task. However, due to page constraints, more experiments are included it in the appendix.

\smallskip\noindent\textbf{Implementation Details.} 
Given the simplicity and high efficiency of the point-based RandLA-Net \cite{RandLA-Net}, we have selected it as the baseline model to evaluate the effectiveness of our U-Next architecture. It is important to note that our architecture is not constrained to RandLA-Net, as demonstrated in Section \ref{sec:ablation}. Specifically, the local feature aggregation module used in RandLA-Net is integrated as coding blocks for U-Next as shown in Equation 1 and Equation 2. During network training, the number of points fed into the network is 40960, the batch size is set to 4 and the maximum training epochs are set to 100. Adam optimizer \cite{Adam} is used with default parameters to optimize the loss function. We follow RandLA-Net \cite{RandLA-Net} to set K=16 for KNN, and the initial learning rate is set to 0.01. All experiments are conducted on a single NVIDIA GeForce RTX3090 GPU and an i7-12700K CPU.  We utilize simple random down-sampling to improve computational efficiency during training.

\subsection{Semantic Segmentation Results}

\smallskip\noindent\textbf{Evaluation on S3DIS.} The Stanford Large-Scale 3D Indoor Spaces (S3DIS) \cite{s3dis} dataset was collected using a Matterport scanner in 6 large indoor areas of 3 different buildings with a total of 272 rooms. Points in this dataset are labeled into 13 semantic categories, and each point is represented as a 6D vector (coordinates ($xyz$) and color ($rgb$)). We follow the widely-used 6-fold cross-validation to comprehensively evaluate the segmentation performance. 

Table~\ref{tab:s3dis} shows the quantitative results achieved by the proposed method and other baselines on the S3DIS dataset. It can be seen that RandLA-Net equipped with the proposed U-Next architecture achieves strong segmentation performance with an overall accuracy of 89.5\% and mIoU score of 73.2\%, outperforming the baseline RandLA-Net by 1.5\% and 3.2\% in OA and mIoU scores. It is also noted that the proposed architecture consistently improves the segmentation performance of all categories, except for class \textit{chairs} with a small fluctuation. In particular, the performance of our method is superior to the baseline in large plane categories such as \textit{walls}, and small instance categories such as \textit{boards}, \textit{columns}, and \textit{beams}. This clearly shows that stacking more sub-networks to capture local geometric details can effectively improve the segmentation performance.

Despite the slight performance gap between our proposed method and SoTA approaches, it is important to reiterate that the primary objective of this paper is not to establish new performance benchmarks. Rather, our goal is to introduce a simple and flexible framework for semantic segmentation that can be readily implemented and expanded upon by other researchers and practitioners.

\smallskip\noindent\textbf{Evaluation on Toronto3D.}
The Toronto3D dataset \cite{Toronto3D} was obtained using a vehicle-mounted MLS system, covering around 1 km of roadway scenarios on Avenue Road in Toronto, Canada. The dataset is divided into four tiles, with each tile covering approximately 250 m, and each point is annotated with one of 8 semantic labels. During training, both 3D coordinates and color attributes are available to use. Following \cite{Toronto3D}, Area 2 is used as the test set, and the remaining three areas are used as training sets.

The quantitative results achieved by our method and other baselines on the Toronto3D dataset are shown in Table~\ref{tab:toronto3D}. Considering several baselines do not use color information as input on this dataset, we thereby report the performance of our method with/without utilizing color information, for a fair comparison. It can be seen that the proposed RandLA-Net (U-Next) achieves the best performance with/without color attributes. In particular, the mIoU scores of the proposed method improve by 1.5\% and 2.2\% respectively, compared with the baseline RandLA-Net.

\begin{table}[]
\centering
\setlength{\abovecaptionskip}{0cm}
\caption{Quantitative results of different approaches on the Toronto3D dataset \cite{Toronto3D} (\textit{Area 2}).}
\scalebox{0.9}{
\begin{tabular}{crcc}
\bottomrule
\multirow{2}*{RGB}&\multirow{2}*{Method}& \multirow{2}*{OA}& \multirow{2}*{mIoU}\\
&&&\\
\hline
\multirow{9}*{No}&PointNet++ \cite{PointNet++} & 92.6 & 59.5\\
~&DGCNN \cite{DGCNN} & 94.2 & 61.7\\
~&MS-PCNN \cite{MS-PCNN} & 90.0 & 65.9\\
~&KPConv \cite{KPConv} & 95.4 & 69.1\\
~&TGNet \cite{TGNet} & 94.1 & 61.3\\
~&MS-TGNet \cite{Toronto3D} & 95.7 & 70.5\\
~&LACV-Net \cite{LACV-Net} & 95.8 & 78.5\\
\cline{2-4}
~&RandLA-Net \cite{RandLA-Net} & 93.0 & 77.7\\
~&\textbf{\textcolor{red}{+ U-Next}} & \textbf{96.0} \textbf{\textcolor{red}{+ 3.0}} & \textbf{79.2} \textbf{\textcolor{red}{+ 1.5}}\\
\hline
\multirow{6}*{Yes}&ResDLPS-Net \cite{ResDLPS-Net} & 96.5 & 80.3\\
~&BAF-LAC \cite{BAF-LAC} & 95.2 & 82.2\\
~&BAAF-Net \cite{BAAF-Net} & 94.2 & 81.2\\
~&LACV-Net \cite{LACV-Net} & 97.4 & 82.7\\
\cline{2-4}
~&RandLA-Net \cite{RandLA-Net} & 94.4 & 81.8\\
~&\textbf{\textcolor{red}{+ U-Next}} & \textbf{97.7} \textbf{\textcolor{red}{+ 3.3}}& \textbf{84.0} \textbf{\textcolor{red}{+ 2.2}}\\
\bottomrule
\end{tabular}}
\label{tab:toronto3D}
\vspace{-1.0em}
\end{table}

\smallskip\noindent\textbf{Evaluation on SensatUrban.}
The SensatUrban dataset \cite{SensatUrban} is a photogrammetric point cloud dataset that spans across 7.6 km$^2$ of the urban landscape, comprising almost 3 billion points that are densely annotated with 13 semantic categories. We follow the official splits to train our network and evaluate the performance on the online test server.

Table~\ref{tab:sensatUrban} shows the quantitative results achieved by our method and other baselines on the SensatUrban dataset. We can see that the proposed method significantly improves the mIoU score from 52.7\% to 62.8\%, with clear improvements in almost all categories. We also notice that the proposed method can achieve an IoU score of 36.8\% on the challenging \textit{railway} class, primarily because the key of our method is to stack as many U-Net $L^1$ codecs as possible with multi-level deep supervision, hence less information loss and small semantic gap are achieved, further leading to improved segmentation performance, especially for small objects. This can be also verified in the categories of class \textit{wall} and \textit{footpath}.

\begin{table*}[]
\centering
\setlength{\abovecaptionskip}{0cm}
\setlength{\belowcaptionskip}{-0.2cm}
\caption{Quantitative results of different approaches on the SenastUrban dataset \cite{SensatUrban}.}
\resizebox{0.95\textwidth}{!}{
\begin{tabular}{rccccccccccccccc}
\bottomrule
\multirow{2}*{Method}& \multirow{2}*{OA}& \multirow{2}*{mIoU}& \multirow{2}*{ground}& \multirow{2}*{veg.}& \multirow{2}*{buil.}& \multirow{2}*{wall}& \multirow{2}*{bri.}& \multirow{2}*{park.}& \multirow{2}*{rail}& \multirow{2}*{traffic.}& \multirow{2}*{street.}& \multirow{2}*{car}& \multirow{2}*{foot.}& \multirow{2}*{bike}& \multirow{2}*{water}\\
&&&&&&&&&&&&&&\\
\hline
PointNet \cite{PointNet} & 80.8 & 23.7 & 67.9 & 89.5 & 80.1 & 0.0 & 0.0 & 3.9 & 0.0 & 31.6 & 0.0 & 35.1 & 0.0 & 0.0 & 0.0 \\
PointNet++ \cite{PointNet++} & 84.3 & 32.9 & 72.5 & 94.2 & 84.8 & 2.7 & 2.1 & 25.8 & 0.0 & 31.5 & 11.4 & 38.8 & 7.1 & 0.0 & 56.9 \\
TagentConv \cite{TagentConv} & 76.9 & 33.3 & 71.5 & 91.4 & 75.9 & 35.2 & 0.0 & 45.3 & 0.0 & 26.7 & 19.2 & 67.6 & 0.0 & 0.0 & 0.0 \\
SPGraph \cite{SPG} & 85.3 & 37.3 & 69.9 & 94.6 & 88.9 & 32.8 & 12.6 & 15.8 & 15.5 & 30.6 & 22.9 & 56.4 & 0.5 & 0.0 & 44.2 \\
SparseConv \cite{SparseConv} & 88.7 & 42.7 & 74.1 & 97.9 & 94.2 & 63.3 & 7.5 & 24.2 & 0.0 & 30.1 & 34.0 & 74.4 & 0.0 & 0.0 & 54.8 \\
KPConv \cite{KPConv} & 93.2 & 57.6 & 87.1 & \textbf{98.9} & 95.3 & \textbf{74.4} & 28.7 & 41.4 & 0.0 & 55.9 & \textbf{54.4} & \textbf{85.7} & 40.4 & 0.0 & \textbf{86.3} \\
BAF-LAC \cite{BAF-LAC} & 91.5 & 54.1 & 84.4 & 98.4 & 94.1 & 57.2 & 27.6 & 42.5 & 15.0 & 51.6 & 39.5 & 78.1 & 40.1 & 0.0 & 75.2 \\
BAAF-Net \cite{BAAF-Net} & 92.0 & 57.3 & 84.2 & 98.3 & 94.0 & 55.2 & 48.9 & 57.7 & 20.0 & 57.3 & 39.3 & 79.3 & 40.7 & 0.0 & 70.1 \\
LACV-Net \cite{LACV-Net} & \textbf{93.2} & 61.3 & \textbf{85.5} & 98.4 & \textbf{95.6} & 61.9 & \textbf{58.6} & \textbf{64.0} & 28.5 & 62.8 & 45.4 & 81.9 & 42.4 & \textbf{4.8} & 67.7 \\
\hline
RandLA-Net \cite{RandLA-Net}  & 89.8 & 52.7 & 80.1 & 98.1 & 91.6 & 48.9 & 40.6 & 51.6 & 0.0 & 56.7 & 33.2 & 80.1 & 32.6 & 0.0 & 71.3 \\
\textbf{\textcolor{red}{+ U-Next}} & \underline{93.0} \textbf{\textcolor{red}{+ 3.2}} & \textbf{62.8} \textbf{\textcolor{red}{+ 10.1}} & \underline{85.2} & \underline{98.6} & \underline{95.0} & \underline{68.2} & \underline{53.6} & \underline{60.4} & \textbf{36.8} & \textbf{64.0} & \underline{48.9} & \underline{84.9} & \textbf{45.1} & 0.0 & \underline{76.2}\\
\bottomrule
\end{tabular}%
\label{tab:sensatUrban}
}
\end{table*}

\begin{figure*}[h]
\setlength{\abovecaptionskip}{0.2cm}
\setlength{\belowcaptionskip}{-0.1cm}
\centering
\includegraphics[width=0.95\textwidth]{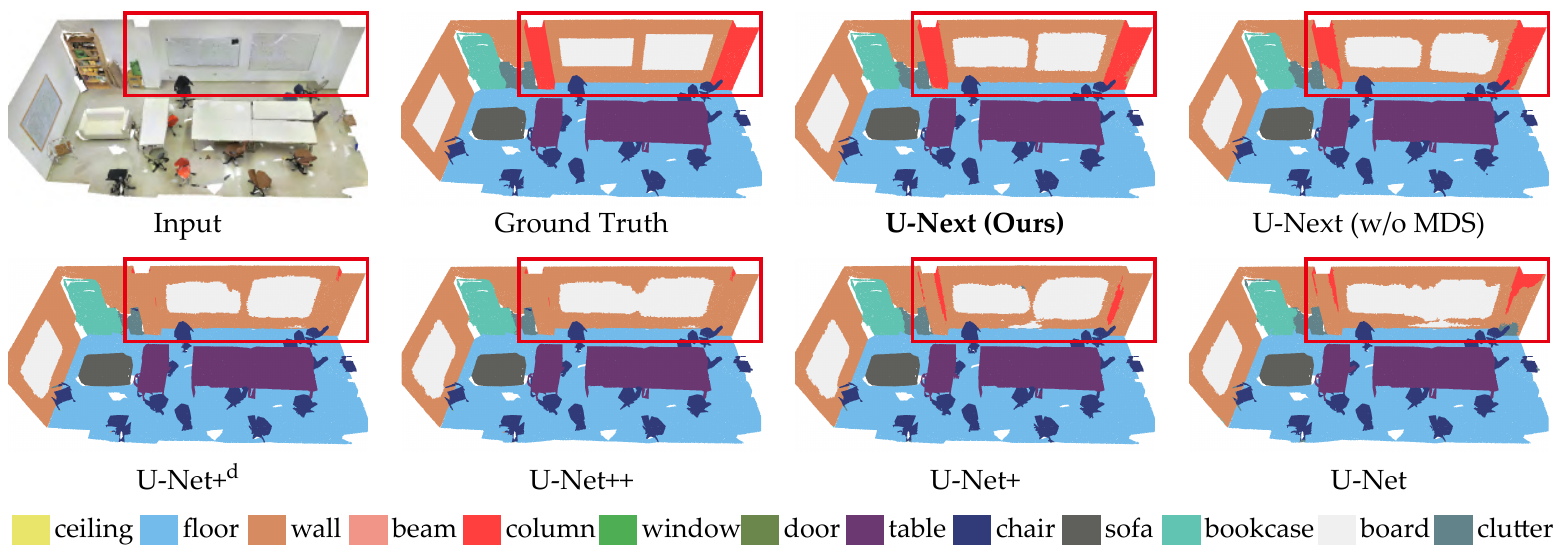}
\caption{Qualitative comparisons of RandLA-Net with different architectures on S3DIS (Area 5). MDS: Multi-level deep supervision.}
\label{fig:arc}
\vspace{-0.5em}
\end{figure*}

\subsection{Ablation Studies}
\label{sec:ablation}

Here, we conduct extensive ablation studies on the proposed U-Next architecture from the following four aspects. All following models are trained on Areas 1-4 and 6, and tested on \textit{Area 5} of the S3DIS \cite{s3dis} dataset.

\vspace{1.5mm}
\noindent \textbf{U-Next with Different Baselines.}
To verify the generalization ability of the proposed U-Next for different baseline models, we incorporate four representative approaches: PointNet++ \cite{PointNet++}, RandLA-Net \cite{RandLA-Net}, BAAF-Net \cite{BAAF-Net} and LACV-Net \cite{LACV-Net}, into our U-Next architecture. For fair comparisons, we only use their local feature extraction modules as coding blocks in Eq.1 and Eq.2, despite several baselines introducing additional loss functions and architectural improvements. Table~\ref{tab:arc} shows the quantitative results achieved by different baselines. It can be seen that the semantic segmentation performance can be consistently improved (with an average improvement of 6.8\% in mIoU scores) after incorporating the baseline model into our U-Next architecture. We also noticed that the improvement for PointNet++ is significant (10.4\%), indicating that our U-Next architecture can significantly improve the feature learning capability of relatively weak backbones, by driving the network to iteratively fuse the multi-scale local feature maps generated by high-to-low and low-to-high sub-networks.

\vspace{1.5mm}
\noindent \textbf{Varying U-Net-like Architectures.} On the other hand, to further demonstrate the effectiveness of the proposed U-Next compared to existing architectures, we take U-Net \cite{UNet}, U-Net+, U-Net++ \cite{UNet++} and the proposed U-Net+$^d$ and U-Next as the framework for semantic segmentation of point clouds, respectively. As shown in the results in Table~\ref{tab:arc},  all baseline methods achieve better performance when using the U-Next architecture, compared with that of other architectures. We also noticed that U-Net++, which is very successful in 2D medical images, failed to significantly improve the semantic segmentation performance of point cloud networks. By contrast, by stacking more U-Net $L^1$ subnetworks and leveraging multi-level deep supervision, the segmentation performance can be steadily improved.

\vspace{1.5mm}
\noindent \textbf{Multi-Level Deep Supervision. }
To further demonstrate the effectiveness of the designed multi-level deep supervision, we further verify the effectiveness of the proposed deep supervision. From the results in Table~\ref{tab:arc}, U-Next with multi-level deep supervision further improves the mIoU score by 1.7\% on average over U-Next without deep supervision. This further shows that stacking U-Net $L^1$ sub-networks and multiple-level deep supervision schemes may complement each other and jointly improve performance.

\vspace{1.5mm}
\noindent \textbf{Computational Costs and Model Complexity.}
To investigate the complexity and computational costs of the proposed \nickname{} architecture, we report the number of parameters and floating point operations (FLOPs) for different architectures with RandLA-Net in Table~\ref{tab:efficient}. Not surprisingly, the proposed U-Next indeed has more trainable parameters and FLOPS, but not significant, since the addition of intermediate nodes does not bring massive trainable parameters, while most of the trainable parameters are generated in the intersection layer of the encoder-decoder architecture (due to the highest dimension in intermediate layers). Considering the segmentation performance and computational cost, the proposed U-Next architecture is meaningful and useful in practice.

Here, we additionally provide the qualitative comparison results achieved by RandLA-Net equipped with different architectures on S3DIS Area 5. As shown in Figure~\ref{fig:arc}, it can be seen that the proposed U-Next can segment the boundary areas smoothly and accurately. Meanwhile, several objects such as columns can be clearly segmented by the proposed U-Next, but not by U-Net or U-Net++, primarily because U-Net and U-Net++ only fuse horizontally with low-to-high feature maps, and cannot effectively explore sufficient multi-scale information for comprehensive representations. More visualizations and experimental results are provided in the supplementary material. 

\begin{table}[h]
\setlength\tabcolsep{3pt}
\vspace{0.3em}
\begin{center}
\setlength{\abovecaptionskip}{0cm}
\caption{Semantic segmentation results (IoU: $\%$) of different baseline models with architectures on S3DIS (Area 5). MDS: Multi-level deep supervision.}
\scalebox{0.8}{
    \begin{tabular}{l|cccc}
    \hline
    \multirow{2}*{\textbf{Architecture}} & \multirow{2}*{\textbf{RandLA-Net}} & \multirow{2}*{\textbf{BAAF-Net}} & \multirow{2}*{\textbf{LACV-Net}} & \multirow{2}*{\textbf{PointNet++}} \\
    &&&&\\
    \hline
    U-Net          & 62.5 & 63.4 & 64.0 & 47.0\\
    U-Net+         & 63.4 \textcolor{red}{+0.9} & 62.9 \textcolor{blue}{-0.5} & 64.4 \textcolor{red}{+0.4} & 49.1 \textcolor{red}{+2.1}\\
    U-Net++        & 63.9 \textcolor{red}{+1.4} & 64.0 \textcolor{red}{+0.6} & 65.1 \textcolor{red}{+1.1} & 50.8 \textcolor{red}{+3.8}\\
    U-Net+$^d$     & 65.9 \textcolor{red}{+3.4} & 65.6 \textcolor{red}{+2.2} & 67.2 \textcolor{red}{+3.2} & 54.9 \textcolor{red}{+7.9}\\
    U-Next w/o MDS  & 66.7 \textcolor{red}{+4.2} & 67.1 \textcolor{red}{+3.7} & 68.0 \textcolor{red}{+4.0} & 55.8 \textcolor{red}{+8.8}\\
    \textbf{U-Next (Ours)}  & 68.3 \textcolor{red}{+5.8} & 68.6 \textcolor{red}{+5.2} & 69.9 \textcolor{red}{+5.9} & 57.4 \textcolor{red}{+10.4}\\
    \hline
    \end{tabular}
    \label{tab:arc}}
\end{center}
\vspace{-0.5em}
\end{table}

\section{Conclusions}
\label{sec:concl}

\begin{table}[]
\begin{center}
\setlength{\abovecaptionskip}{0cm}
\setlength{\belowcaptionskip}{-0.2cm}
\caption{Model parameters (millions) and FLOPs (millions) of RandLA-Net with different architectures.}
\scalebox{0.8}{
    \begin{tabular}{l|cc}
    \hline
    \multirow{2}*{\textbf{Architecture}} & \multirow{2}*{\textbf{Parameters}} & \multirow{2}*{\textbf{FLOPs}}\\
    &&\\
    \hline
    U-Net       & 4.99 & 31.35\\
    U-Net+      & 5.24 & 31.96\\
    U-Net++     & 5.39 & 33.02\\
    U-Net+$^d$  & 5.51 & 33.79\\
    U-Next      & 5.61 & 34.51\\
    \hline
    \end{tabular}
    \label{tab:efficient}}
\end{center}
\vspace{-2.0em}
\end{table}

In this paper, we propose a U-Net-like framework for point cloud semantic segmentation, termed U-Next. The key to our U-Next framework is to stack a maximum number of U-Net $L^1$ sub-networks, ensuring minimal semantic gaps and better multi-scale feature representations. We also introduce a multi-level deep supervision mechanism to facilitate smooth gradient propagation and provide additional guidance at different levels of abstraction. Extensive experimental results on several large-scale benchmarks and different backbones demonstrate the effectiveness and versatility of the proposed framework. In future work, we plan to explore the applicability of U-Next to different modality data and tasks to further validate its generalizability and potential in diverse applications.


{\small
\bibliographystyle{ieee_fullname}
\bibliography{egbib}

\begin{thebibliography}{10}\itemsep=-1pt

\bibitem{ao2022you}
Sheng Ao, Yulan Guo, Qingyong Hu, Bo Yang, Andrew Markham, and Zengping Chen.
\newblock You only train once: Learning general and distinctive {3D} local
  descriptors.
\newblock {\em IEEE Transactions on Pattern Analysis and Machine Intelligence},
  2022.

\bibitem{s3dis}
Iro Armeni, Ozan Sener, Amir~R. Zamir, Helen Jiang, Ioannis Brilakis, Martin
  Fischer, and Silvio Savarese.
\newblock {3D} semantic parsing of large-scale indoor spaces.
\newblock In {\em IEEE/CVF Conference on Computer Vision and Pattern
  Recognition (CVPR)}, 2016.

\bibitem{MF_a_2}
Vijay Badrinarayanan, Alex Kendall, and Roberto Cipolla.
\newblock Segnet: A deep convolutional encoder-decoder architecture for image
  segmentation.
\newblock {\em IEEE Transactions on Pattern Analysis and Machine Intelligence},
  pages 2481--2495, 2017.

\bibitem{SemanticKITTI}
Jens Behley, Martin Garbade, Andres Milioto, Jan Quenzel, Sven Behnke, Cyrill
  Stachniss, and Jurgen Gall.
\newblock {SemanticKITTI: A Dataset for Semantic Scene Understanding of LiDAR
  Sequences}.
\newblock In {\em IEEE/CVF International Conference on Computer Vision (ICCV)},
  2019.

\bibitem{blanc2020genuage}
Thomas Blanc, Mohamed El~Beheiry, Cl{\'e}ment Caporal, Jean-Baptiste Masson,
  and Bassam Hajj.
\newblock Genuage: visualize and analyze multidimensional single-molecule point
  cloud data in virtual reality.
\newblock {\em Nature Methods}, 17(11):1100--1102, 2020.

\bibitem{boulch2017unstructured}
A Boulch, B~Le Saux, and N Audebert.
\newblock Unstructured point cloud semantic labeling using deep segmentation
  networks.
\newblock In {\em Eurographics Workshop on 3D Object Retrieval (3DOR)}, 2017.

\bibitem{MF_1}
Zhaowei Cai, Quanfu Fan, Rogerio~S Feris, and Nuno Vasconcelos.
\newblock A unified multi-scale deep convolutional neural network for fast
  object detection.
\newblock In {\em European Conference on Computer Vision (ECCV)}, 2016.

\bibitem{TransUNet}
Jieneng Chen, Yongyi Lu, Qihang Yu, Xiangde Luo, Ehsan Adeli, Yan Wang, Le Lu,
  Alan~L Yuille, and Yuyin Zhou.
\newblock Transunet: Transformers make strong encoders for medical image
  segmentation.
\newblock {\em arXiv:2102.04306}, 2021.

\bibitem{MF_2}
Liang-Chieh Chen, George Papandreou, Iasonas Kokkinos, Kevin Murphy, and Alan~L
  Yuille.
\newblock Deeplab: Semantic image segmentation with deep convolutional nets,
  atrous convolution, and fully connected crfs.
\newblock {\em IEEE Transactions on Pattern Analysis and Machine Intelligence},
  pages 834--848, 2017.

\bibitem{chen2022stpls3d}
Meida Chen, Qingyong Hu, Thomas Hugues, Andrew Feng, Yu Hou, Kyle McCullough,
  and Lucio Soibelman.
\newblock Stpls{3D}: A large-scale synthetic and real aerial photogrammetry
  {3D} point cloud dataset.
\newblock 2022.

\bibitem{PB_1}
Xiaozhi Chen, Huimin Ma, Ji Wan, Bo Li, and Tian Xia.
\newblock Multi-view {3D} object detection network for autonomous driving.
\newblock In {\em IEEE/CVF Conference on Computer Vision and Pattern
  Recognition (CVPR)}, 2017.

\bibitem{DS_2}
Yu Chen, Chunhua Shen, Xiu-Shen Wei, Lingqiao Liu, and Jian Yang.
\newblock Adversarial posenet: A structure-aware convolutional network for
  human pose estimation.
\newblock In {\em IEEE/CVF International Conference on Computer Vision (ICCV)},
  2017.

\bibitem{4dMinkpwski}
Christopher Choy, JunYoung Gwak, and Silvio Savarese.
\newblock 4{D} spatio-temporal convnets: Minkowski convolutional neural
  networks.
\newblock In {\em IEEE/CVF Conference on Computer Vision and Pattern
  Recognition (CVPR)}, 2019.

\bibitem{Dou}
Qi Dou, Hao Chen, Yueming Jin, Lequan Yu, Jing Qin, and Pheng-Ann Heng.
\newblock {3D} deeply supervised network for automatic liver segmentation from
  ct volumes.
\newblock In {\em International Conference on Medical Image Computing and
  Computer-Assisted Intervention}, 2016.

\bibitem{ResDLPS-Net}
Jing Du, Guorong Cai, Zongyue Wang, Shangfeng Huang, Jinhe Su, José {Marcato
  Junior}, Julian Smit, and Jonathan Li.
\newblock Resdlps-net: Joint residual-dense optimization for large-scale point
  cloud semantic segmentation.
\newblock {\em ISPRS Journal of Photogrammetry and Remote Sensing}, 2021.

\bibitem{SCF-Net}
Siqi Fan, Qiulei Dong, Fenghua Zhu, Yisheng Lv, Peijun Ye, and Fei-Yue Wang.
\newblock Scf-net: Learning spatial contextual features for large-scale point
  cloud segmentation.
\newblock In {\em IEEE/CVF Conference on Computer Vision and Pattern
  Recognition (CVPR)}, 2021.

\bibitem{SparseConv}
Benjamin Graham, Martin Engelcke, and Laurens Van Der~Maaten.
\newblock {3D} semantic segmentation with submanifold sparse convolutional
  networks.
\newblock In {\em IEEE/CVF Conference on Computer Vision and Pattern
  Recognition (CVPR)}, 2018.

\bibitem{VB_2}
Benjamin Graham, Martin Engelcke, and Laurens Van Der~Maaten.
\newblock {3D} semantic segmentation with submanifold sparse convolutional
  networks.
\newblock In {\em IEEE/CVF Conference on Computer Vision and Pattern
  Recognition (CVPR)}, 2018.

\bibitem{DenseUNet}
Steven Guan, Amir~A Khan, Siddhartha Sikdar, and Parag~V Chitnis.
\newblock Fully dense unet for {2-D} sparse photoacoustic tomography artifact
  removal.
\newblock {\em IEEE Journal of Biomedical and Health Informatics}, pages
  568--576, 2019.

\bibitem{PCT}
Meng-Hao Guo, Jun-Xiong Cai, Zheng-Ning Liu, Tai-Jiang Mu, Ralph~R Martin, and
  Shi-Min Hu.
\newblock Pct: Point cloud transformer.
\newblock {\em Computational Visual Media}, 2021.

\bibitem{guo2020deep}
Yulan Guo, Hanyun Wang, Qingyong Hu, Hao Liu, Li Liu, and Mohammed Bennamoun.
\newblock Deep learning for {3D} point clouds: A survey.
\newblock {\em IEEE Transactions on Pattern Analysis and Machine Intelligence},
  43(12):4338--4364, 2020.

\bibitem{Semantic3D}
Timo Hackel, N. Savinov, L. Ladicky, Jan~D. Wegner, K. Schindler, and M.
  Pollefeys.
\newblock {SEMANTIC{3D}.NET: A new large-scale point cloud classification
  benchmark}.
\newblock In {\em ISPRS Annals of the Photogrammetry, Remote Sensing and
  Spatial Information Sciences}, 2017.

\bibitem{PT}
Zhao Hengshuang, Jiang Li, Jia Jiaya, Torr~Philip H.S., and Koltun Vladlen.
\newblock Point transformer.
\newblock In {\em IEEE/CVF International Conference on Computer Vision (ICCV)},
  2021.

\bibitem{hu2022sqn}
Qingyong Hu, Bo Yang, Guangchi Fang, Yulan Guo, Ale{\v{s}} Leonardis, Niki
  Trigoni, and Andrew Markham.
\newblock Sqn: Weakly-supervised semantic segmentation of large-scale {3D}
  point clouds.
\newblock In {\em European Conference on Computer Vision (ECCV)}, pages
  600--619. Springer, 2022.

\bibitem{SensatUrban}
Qingyong Hu, Bo Yang, Sheikh Khalid, Wen Xiao, Niki Trigoni, and Andrew
  Markham.
\newblock Towards semantic segmentation of urban-scale {3D} point clouds: A
  dataset, benchmarks and challenges.
\newblock In {\em IEEE/CVF Conference on Computer Vision and Pattern
  Recognition (CVPR)}, 2021.

\bibitem{hu2022sensaturban}
Qingyong Hu, Bo Yang, Sheikh Khalid, Wen Xiao, Niki Trigoni, and Andrew
  Markham.
\newblock Sensaturban: Learning semantics from urban-scale photogrammetric
  point clouds.
\newblock {\em International Journal of Computer Vision}, 130(2):316--343,
  2022.

\bibitem{RandLA-Net}
Qingyong Hu, Bo Yang, Linhai Xie, Stefano Rosa, Yulan Guo, Zhihua Wang, Niki
  Trigoni, and Andrew Markham.
\newblock Randla-net: Efficient semantic segmentation of large-scale point
  clouds.
\newblock In {\em IEEE/CVF Conference on Computer Vision and Pattern
  Recognition (CVPR)}, 2020.

\bibitem{hu2021learning}
Qingyong Hu, Bo Yang, Linhai Xie, Stefano Rosa, Yulan Guo, Zhihua Wang, Niki
  Trigoni, and Andrew Markham.
\newblock Learning semantic segmentation of large-scale point clouds with
  random sampling.
\newblock {\em IEEE Transactions on Pattern Analysis and Machine Intelligence},
  2021.

\bibitem{UNet3}
Huimin Huang, Lanfen Lin, Ruofeng Tong, Hongjie Hu, Qiaowei Zhang, Yutaro
  Iwamoto, Xianhua Han, Yen-Wei Chen, and Jian Wu.
\newblock Unet 3+: A full-scale connected unet for medical image segmentation.
\newblock In {\em IEEE International Conference on Acoustics, Speech and Signal
  Processing}, 2020.

\bibitem{RSNet}
Qiangui Huang, Weiyue Wang, and Ulrich Neumann.
\newblock Recurrent slice networks for {3D} segmentation of point clouds.
\newblock In {\em IEEE/CVF Conference on Computer Vision and Pattern
  Recognition (CVPR)}, 2018.

\bibitem{Adam}
Diederik~P Kingma and Jimmy Ba.
\newblock Adam: A method for stochastic optimization.
\newblock {\em International Conference on Learning Representations}, 2015.

\bibitem{kundu2020virtual}
Abhijit Kundu, Xiaoqi Yin, Alireza Fathi, David Ross, Brian Brewington, Thomas
  Funkhouser, and Caroline Pantofaru.
\newblock Virtual multi-view fusion for 3{D} semantic segmentation.
\newblock In {\em European Conference on Computer Vision (ECCV)}, 2020.

\bibitem{tradition_1}
Loic Landrieu, Hugo Raguet, Bruno Vallet, Cl{\'e}ment Mallet, and Martin
  Weinmann.
\newblock A structured regularization framework for spatially smoothing
  semantic labelings of {3D} point clouds.
\newblock {\em ISPRS Journal of Photogrammetry and Remote Sensing}, 2017.

\bibitem{SPG}
Loic Landrieu and Martin Simonovsky.
\newblock Large-scale point cloud semantic segmentation with superpoint graphs.
\newblock In {\em IEEE/CVF Conference on Computer Vision and Pattern
  Recognition (CVPR)}, 2018.

\bibitem{PB_2}
Alex~H Lang, Sourabh Vora, Holger Caesar, Lubing Zhou, Jiong Yang, and Oscar
  Beijbom.
\newblock Pointpillars: Fast encoders for object detection from point clouds.
\newblock In {\em IEEE/CVF Conference on Computer Vision and Pattern
  Recognition (CVPR)}, 2019.

\bibitem{VB_1}
Truc Le and Ye Duan.
\newblock Pointgrid: A deep network for {3D} shape understanding.
\newblock In {\em IEEE/CVF Conference on Computer Vision and Pattern
  Recognition (CVPR)}, 2018.

\bibitem{Lee}
Chen-Yu Lee, Saining Xie, Patrick Gallagher, Zhengyou Zhang, and Zhuowen Tu.
\newblock Deeply-supervised nets.
\newblock In {\em Artificial Intelligence and Statistics}, 2015.

\bibitem{DeepGCNs}
Guohao Li, Matthias Muller, Ali Thabet, and Bernard Ghanem.
\newblock Deepgcns: Can gcns go as deep as cnns?
\newblock In {\em IEEE/CVF International Conference on Computer Vision (ICCV)},
  2019.

\bibitem{TGNet}
Ying Li, Lingfei Ma, Zilong Zhong, Dongpu Cao, and Jonathan Li.
\newblock {TGNet: Geometric Graph CNN on {3-D} Point Cloud Segmentation}.
\newblock In {\em IEEE Transactions on Geoscience and Remote Sensing}, 2020.

\bibitem{Point-Voxel_CNN}
Zhijian Liu, Haotian Tang, Yujun Lin, and Song Han.
\newblock Point-voxel cnn for efficient 3{D} deep learning.
\newblock In {\em Advances in Neural Information Processing Systems (NeurIPS)},
  2019.

\bibitem{MS-PCNN}
Lingfei Ma, Ying Li, Jonathan Li, Weikai Tan, Yongtao Yu, and Michael~A
  Chapman.
\newblock {Multi-Scale Point-Wise Convolutional Neural Networks for {3D} Object
  Segmentation From LiDAR Point Clouds in Large-Scale Environments}.
\newblock In {\em IEEE Transactions on Intelligent Transportation Systems},
  2021.

\bibitem{VB_3}
Daniel Maturana and Sebastian Scherer.
\newblock Voxnet: A {3D} convolutional neural network for real-time object
  recognition.
\newblock In {\em IEEE/RSJ International Conference on Intelligent Robots and
  Systems}, 2015.

\bibitem{MF_a_1}
Alejandro Newell, Kaiyu Yang, and Jia Deng.
\newblock Stacked hourglass networks for human pose estimation.
\newblock In {\em European Conference on Computer Vision (ECCV)}, 2016.

\bibitem{KPpyramid}
Dong Nie, Rui Lan, Ling Wang, and Xiaofeng Ren.
\newblock Pyramid architecture for multi-scale processing in point cloud
  segmentation.
\newblock In {\em IEEE/CVF Conference on Computer Vision and Pattern
  Recognition (CVPR)}, 2022.

\bibitem{PointNet}
Charles~Ruizhongtai Qi, Hao Su, Kaichun Mo, and Leonidas~J Guibas.
\newblock Pointnet: Deep learning on point sets for {3D} classification and
  segmentation.
\newblock In {\em IEEE/CVF Conference on Computer Vision and Pattern
  Recognition (CVPR)}, 2017.

\bibitem{PointNet++}
Charles~Ruizhongtai Qi, Li Yi, Hao Su, and Leonidas~J Guibas.
\newblock Pointnet++: Deep hierarchical feature learning on point sets in a
  metric space.
\newblock In {\em Advances in Neural Information Processing Systems (NeurIPS)}.

\bibitem{PointNext}
Guocheng Qian, Yuchen Li, Houwen Peng, Jinjie Mai, Hasan Abed Al~Kader Hammoud,
  Mohamed Elhoseiny, and Bernard Ghanem.
\newblock Pointnext: Revisiting pointnet++ with improved training and scaling
  strategies.
\newblock {\em Advances in Neural Information Processing Systems (NeurIPS)},
  2022.

\bibitem{PViT}
Guocheng Qian, Xingdi Zhang, Abdullah Hamdi, and Bernard Ghanem.
\newblock Improving standard transformer models for 3d point cloud
  understanding with image pretraining.
\newblock {\em arXiv preprint arXiv:2208.12259}, 2022.

\bibitem{BAAF-Net}
Shi Qiu, Saeed Anwar, and Nick Barnes.
\newblock Semantic segmentation for real point cloud scenes via bilateral
  augmentation and adaptive fusion.
\newblock In {\em IEEE/CVF Conference on Computer Vision and Pattern
  Recognition (CVPR)}, 2021.

\bibitem{UNet}
Olaf Ronneberger, Philipp Fischer, and Thomas Brox.
\newblock U-net: Convolutional networks for biomedical image segmentation.
\newblock In {\em International Conference on Medical Image Computing and
  Computer-Assisted Intervention}, 2015.

\bibitem{tradition_2}
Radu~Bogdan Rusu, Nico Blodow, and Michael Beetz.
\newblock Fast point feature histograms (fpfh) for {3D} registration.
\newblock In {\em IEEE International Conference on Robotics and Automation},
  2009.

\bibitem{pcl}
Radu~Bogdan Rusu and Steve Cousins.
\newblock {3D} is here: Point cloud library (pcl).
\newblock In {\em IEEE International Conference on Robotics and Automation}.
  IEEE, 2011.

\bibitem{MF_3}
Mohamed Samy, Karim Amer, Kareem Eissa, Mahmoud Shaker, and Mohamed ElHelw.
\newblock Nu-net: Deep residual wide field of view convolutional neural network
  for semantic segmentation.
\newblock In {\em IEEE/CVF Conference on Computer Vision and Pattern
  Recognition Workshops (CVPRW)}, 2018.

\bibitem{MF_4}
Shreyas Saxena and Jakob Verbeek.
\newblock Convolutional neural fabrics.
\newblock {\em Advances in Neural Information Processing Systems (NeurIPS)},
  2016.

\bibitem{BAF-LAC}
Hui Shuai, Xiang Xu, and Qingshan Liu.
\newblock Backward attentive fusing network with local aggregation classifier
  for {3D} point cloud semantic segmentation.
\newblock {\em IEEE Transactions on Image Processing}, 2021.

\bibitem{HRNet}
Ke Sun, Bin Xiao, Dong Liu, and Jingdong Wang.
\newblock Deep high-resolution representation learning for human pose
  estimation.
\newblock In {\em IEEE/CVF Conference on Computer Vision and Pattern
  Recognition (CVPR)}, 2019.

\bibitem{GoogLeNet}
Christian Szegedy, Wei Liu, Yangqing Jia, Pierre Sermanet, Scott Reed, Dragomir
  Anguelov, Dumitru Erhan, Vincent Vanhoucke, and Andrew Rabinovich.
\newblock Going deeper with convolutions.
\newblock In {\em IEEE/CVF Conference on Computer Vision and Pattern
  Recognition (CVPR)}, 2015.

\bibitem{Toronto3D}
Weikai Tan, Nannan Qin, Lingfei Ma, Ying Li, Jing Du, Guorong Cai, Ke Yang, and
  Jonathan Li.
\newblock {Toronto-{3D}: A Large-scale Mobile LiDAR Dataset for Semantic
  Segmentation of Urban Roadways}.
\newblock In {\em IEEE/CVF Conference on Computer Vision and Pattern
  Recognition Workshops (CVPRW)}, 2020.

\bibitem{e3d}
Haotian Tang, Zhijian Liu, Shengyu Zhao, Yujun Lin, Ji Lin, Hanrui Wang, and
  Song Han.
\newblock Searching efficient 3{D} architectures with sparse point-voxel
  convolution.
\newblock In {\em European Conference on Computer Vision (ECCV)}, 2020.

\bibitem{CBL}
Liyao Tang, Yibing Zhan, Zhe Chen, Baosheng Yu, and Dacheng Tao.
\newblock Contrastive boundary learning for point cloud segmentation.
\newblock In {\em IEEE/CVF Conference on Computer Vision and Pattern
  Recognition (CVPR)}, 2022.

\bibitem{TagentConv}
Maxim Tatarchenko, Jaesik Park, Vladlen Koltun, and Qian-Yi Zhou.
\newblock Tangent convolutions for dense prediction in {3D}.
\newblock In {\em IEEE/CVF Conference on Computer Vision and Pattern
  Recognition (CVPR)}, 2018.

\bibitem{tchapmi2017segcloud}
Lyne Tchapmi, Christopher Choy, Iro Armeni, JunYoung Gwak, and Silvio Savarese.
\newblock Segcloud: Semantic segmentation of {3D} point clouds.
\newblock In {\em International Conference on 3D Vision}, 2017.

\bibitem{KPConv}
Hugues Thomas, Charles~R. Qi, Jean-Emmanuel Deschaud, Beatriz Marcotegui,
  François Goulette, and Leonidas Guibas.
\newblock Kpconv: Flexible and deformable convolution for point clouds.
\newblock In {\em IEEE/CVF International Conference on Computer Vision (ICCV)},
  2019.

\bibitem{DS_3}
Jonathan Tompson, Ross Goroshin, Arjun Jain, Yann LeCun, and Christoph Bregler.
\newblock Efficient object localization using convolutional networks.
\newblock In {\em IEEE/CVF International Conference on Computer Vision (ICCV)},
  2015.

\bibitem{InsEmb}
He Tong, Liu Yifan, Shen Chunhua, Wang Xinlong, and Sun Changming.
\newblock Instance-aware embedding for point cloud instance segmentation.
\newblock In {\em European Conference on Computer Vision (ECCV)}, 2020.

\bibitem{UCTransNet}
Haonan Wang, Peng Cao, Jiaqi Wang, and Osmar~R Zaiane.
\newblock Uctransnet: rethinking the skip connections in u-net from a
  channel-wise perspective with transformer.
\newblock In {\em Proceedings of the AAAI Conference on Artificial
  Intelligence}, 2022.

\bibitem{MF_a_4}
Jingdong Wang, Zhen Wei, Ting Zhang, and Wenjun Zeng.
\newblock Deeply-fused nets.
\newblock {\em arXiv:1605.07716}, 2016.

\bibitem{GACNet}
Lei Wang, Yuchun Huang, Yaolin Hou, Shenman Zhang, and Jie Shan.
\newblock Graph attention convolution for point cloud semantic segmentation.
\newblock In {\em IEEE/CVF Conference on Computer Vision and Pattern
  Recognition (CVPR)}, 2019.

\bibitem{DGCNN}
Yue Wang, Yongbin Sun, Ziwei Liu, Sanjay~E Sarma, Michael~M Bronstein, and
  Justin~M Solomon.
\newblock Dynamic graph cnn for learning on point clouds.
\newblock {\em ACM Transactions On Graphics}, 2019.

\bibitem{AD-SAGC}
Chi-Chong Wong and Chi-Man Vong.
\newblock Efficient outdoor {3D} point cloud semantic segmentation for critical
  road objects and distributed contexts.
\newblock In {\em European Conference on Computer Vision (ECCV)}, 2020.

\bibitem{PointConv}
Wenxuan Wu, Zhongang Qi, and Li Fuxin.
\newblock Pointconv: Deep convolutional networks on 3d point clouds.
\newblock In {\em IEEE/CVF Conference on Computer Vision and Pattern
  Recognition (CVPR)}, 2019.

\bibitem{ResUNet}
Xiao Xiao, Shen Lian, Zhiming Luo, and Shaozi Li.
\newblock Weighted res-unet for high-quality retina vessel segmentation.
\newblock In {\em International Conference on Information Technology in
  Medicine and Education}, 2018.

\bibitem{ASIS}
Wang Xinlong, Liu Shu, Shen Xiaoyong, Shen Chunhua, and Jia Jiaya.
\newblock Associatively segmenting instances and semantics in point clouds.
\newblock In {\em IEEE/CVF Conference on Computer Vision and Pattern
  Recognition (CVPR)}, 2019.

\bibitem{PointASNL}
Xu Yan, Chaoda Zheng, Zhen Li, Sheng Wang, and Shuguang Cui.
\newblock Pointasnl: Robust point clouds processing using nonlocal neural
  networks with adaptive sampling.
\newblock In {\em IEEE/CVF Conference on Computer Vision and Pattern
  Recognition (CVPR)}, 2020.

\bibitem{3D-BoNet}
Bo Yang, Jianan Wang, Ronald Clark, Qingyong Hu, Sen Wang, Andrew Markham, and
  Niki Trigoni.
\newblock Learning object bounding boxes for 3d instance segmentation on point
  clouds.
\newblock In {\em Advances in Neural Information Processing Systems (NeurIPS)},
  2019.

\bibitem{3P-RNN}
Xiaoqing Ye, Jiamao Li, Hexiao Huang, Liang Du, and Xiaolin Zhang.
\newblock {3D} recurrent neural networks with context fusion for point cloud
  semantic segmentation.
\newblock In {\em European Conference on Computer Vision (ECCV)}, 2018.

\bibitem{PB_3}
Tan Yu, Jingjing Meng, and Junsong Yuan.
\newblock Multi-view harmonized bilinear network for {3D} object recognition.
\newblock In {\em IEEE/CVF Conference on Computer Vision and Pattern
  Recognition (CVPR)}, 2018.

\bibitem{LACV-Net}
Ziyin Zeng, Yongyang Xu, Zhong Xie, Wei Tang, Jie Wan, and Weichao Wu.
\newblock Lacv-net: Semantic segmentation of large-scale point cloud scene via
  local adaptive and comprehensive vlad.
\newblock {\em arXiv:2210.05870}, 2022.

\bibitem{LEARD}
Ziyin Zeng, Yongyang Xu, Zhong Xie, Wei Tang, Jie Wan, and Weichao Wu.
\newblock Leard-net: Semantic segmentation for large-scale point cloud scene.
\newblock {\em International Journal of Applied Earth Observation and
  Geoinformation}, 2022.

\bibitem{ShellNet}
Zhiyuan Zhang, Binh-Son Hua, and Sai-Kit Yeung.
\newblock Shellnet: Efficient point cloud convolutional neural networks using
  concentric shells statistics.
\newblock In {\em IEEE/CVF International Conference on Computer Vision (ICCV)},
  2019.

\bibitem{PointWeb}
Hengshuang Zhao, Li Jiang, Chi-Wing Fu, and Jiaya Jia.
\newblock Pointweb: Enhancing local neighborhood features for point cloud
  processing.
\newblock In {\em IEEE/CVF Conference on Computer Vision and Pattern
  Recognition (CVPR)}, 2019.

\bibitem{MF_a_3}
Hengshuang Zhao, Jianping Shi, Xiaojuan Qi, Xiaogang Wang, and Jiaya Jia.
\newblock Pyramid scene parsing network.
\newblock In {\em IEEE/CVF Conference on Computer Vision and Pattern
  Recognition (CVPR)}, 2017.

\bibitem{PN++}
Liang Zhihao, Li Zhihao, Xu Songcen, Tan Mingkui, and Jia Kui.
\newblock Self-prediction for joint instance and semantic segmentation of point
  clouds.
\newblock In {\em European Conference on Computer Vision (ECCV)}, 2020.

\bibitem{UNet++}
Zongwei Zhou, Md~Mahfuzur~Rahman Siddiquee, Nima Tajbakhsh, and Jianming Liang.
\newblock Unet++: Redesigning skip connections to exploit multiscale features
  in image segmentation.
\newblock {\em IEEE Transactions on Medical Imaging}, pages 1856--1867, 2019.

\bibitem{DS_1}
Qikui Zhu, Bo Du, Baris Turkbey, Peter~L Choyke, and Pingkun Yan.
\newblock Deeply-supervised cnn for prostate segmentation.
\newblock In {\em International Joint Conference on Neural Networks}, 2017.

\end{thebibliography}
}
\clearpage
\appendix
\centerline{\textbf{\Huge{Appendix}}}

\vspace{6mm}
\centerline{\textbf{\large{Overview}}}

\vspace{2mm}
\noindent This supplementary material is organized as follows:
\begin{itemize}
\item Section~\ref{sec:exp_de} details the network architectures.
\item Section~\ref{sec:exp_re} presents the evaluation metrics used in our experiments and provides additional quantitative segmentation results on various benchmarks and tasks. Additionally, we conduct ablation studies to analyze the effectiveness of our proposed approach.
\item Section~\ref{sec:vis} shows more visualizations of semantic segmentation results.
\end{itemize}

\section{Details of the Network Architecture}
\label{sec:exp_de}

In Figure~\ref{fig:unext} of the main paper, we present the general architecture of U-Next. In this section, we provide more details of the proposed architecture.

First, a point cloud with dimension $N\times d_{in}$ is input to the network (the batch dimension is removed for simplicity), where $N$ is the number of points in point cloud and $d_{in}$ is the feature dimension of input points. For both S3DIS \cite{s3dis}, Toronto3D \cite{Toronto3D}, and SensatUrban \cite{SensatUrban} datasets, each point is represented by its 3D coordinates and color information (\textit{i.e.,} x-y-z-R-G-B). We apply a single-layer MLP containing eight $1\times 1$ kernels with ReLU and BN to extract the initial embedding.

The network gradually reduces the resolutions of the point cloud from high to low: ($\frac{N}{4}\rightarrow \frac{N}{16}\rightarrow \frac{N}{64}\rightarrow \frac{N}{256}\rightarrow \frac{N}{512}$), whereas the channel dimension of the features increases as: ($16\rightarrow 64\rightarrow 128\rightarrow 256\rightarrow 512$), as shown in Table~\ref{tab:output}. Note that the number of points and feature dimensions are the output of each coding block, and in the same layer (same \textit{i}), the output of each coding block has the same number of points and the same dimension of features. In the process, point clouds with different resolutions are generated, and the proposed U-Next utilizes multiple sub-networks to fuse the multi-scale feature maps. 

Finally, we use three fully-connected layers with a dropout layer ($64\rightarrow 32\rightarrow (\mathrm{dp}) \rightarrow c$, the drop-out ratio is set as 0.5), and a softmax layer to convert the abstract semantic features into label classification scores for each point in the point cloud. The semantic segmentation result is determined by the label with the highest score.

\begin{figure*}
\vspace{5.0em}
\centering
\includegraphics[width=1.0\textwidth]{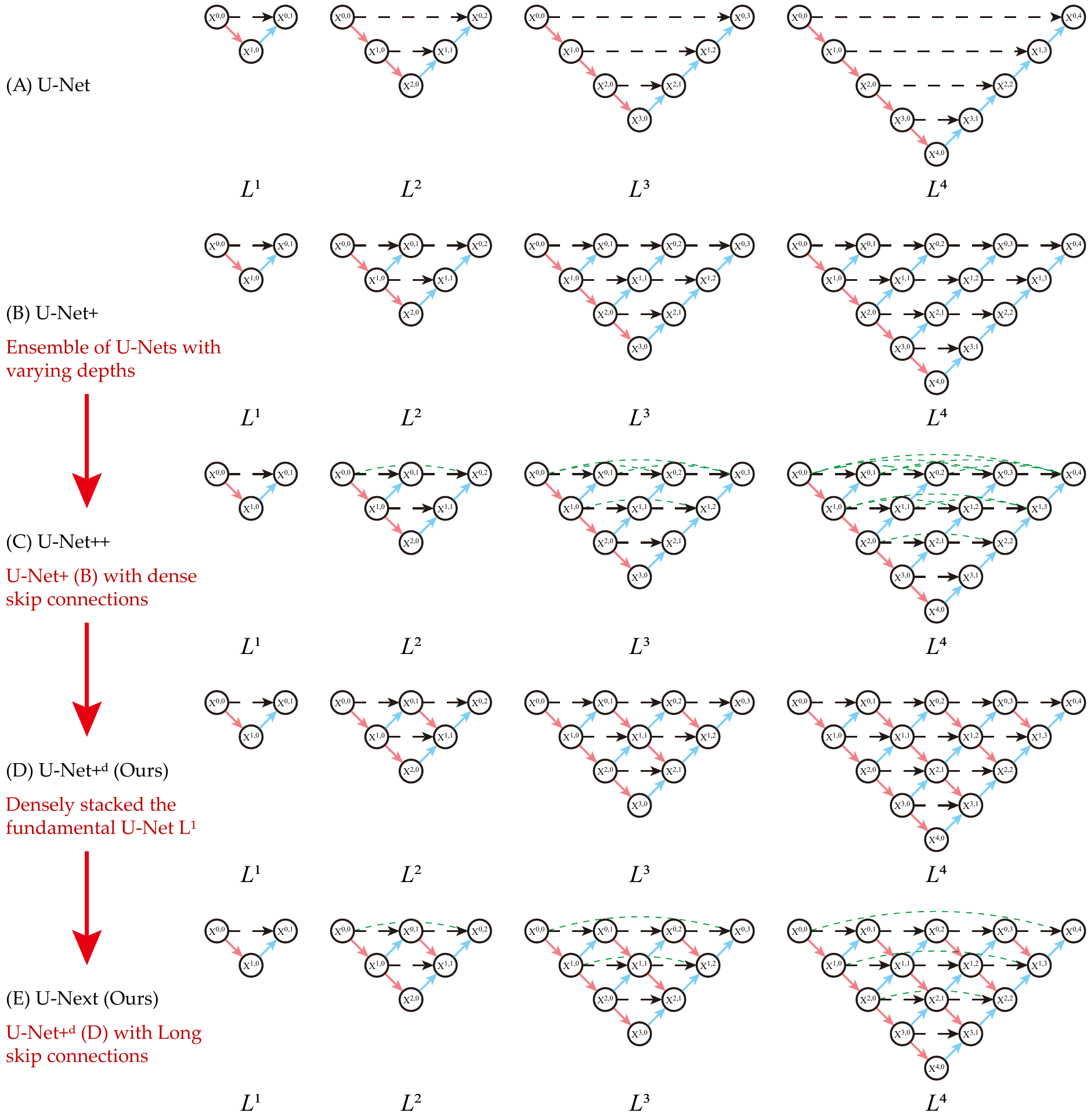}
\caption{Detailed level evolution of U-Net variants.}
\label{fig:evo_de}
\vspace{5.0em}
\end{figure*}

\vspace{1.5mm}
\noindent \textbf{Evolution details of the U-Net variants. } As shown in Figure~\ref{fig:evo_de}, we further provide a comparison of the evolution details of the U-Net variants, including U-Net \cite{UNet}, U-Net+, U-Net++ \cite{UNet++}, U-Net+$^d$ and the proposed U-Next architectures. It is clear that U-Net+ and U-Net++ are fundamentally an ensemble of different levels of U-Net, but with shared encoding layers. Similarly, the proposed U-Next can also be seen as an integration of U-Net, but entirely consists of U-Net $L^1$ sub-networks, which means each node in U-Next incorporates horizontal, low-to-high and high-to-low features, while also with minimal semantic gaps. Additionally, different from U-Net++, our U-Next does not use dense skip connections, primarily because of the efficiency consideration (validated in Section~\ref{sec:abla}). 

\begin{figure*}
\setlength{\abovecaptionskip}{0.1cm}
\centering
\includegraphics[width=1.0\textwidth]{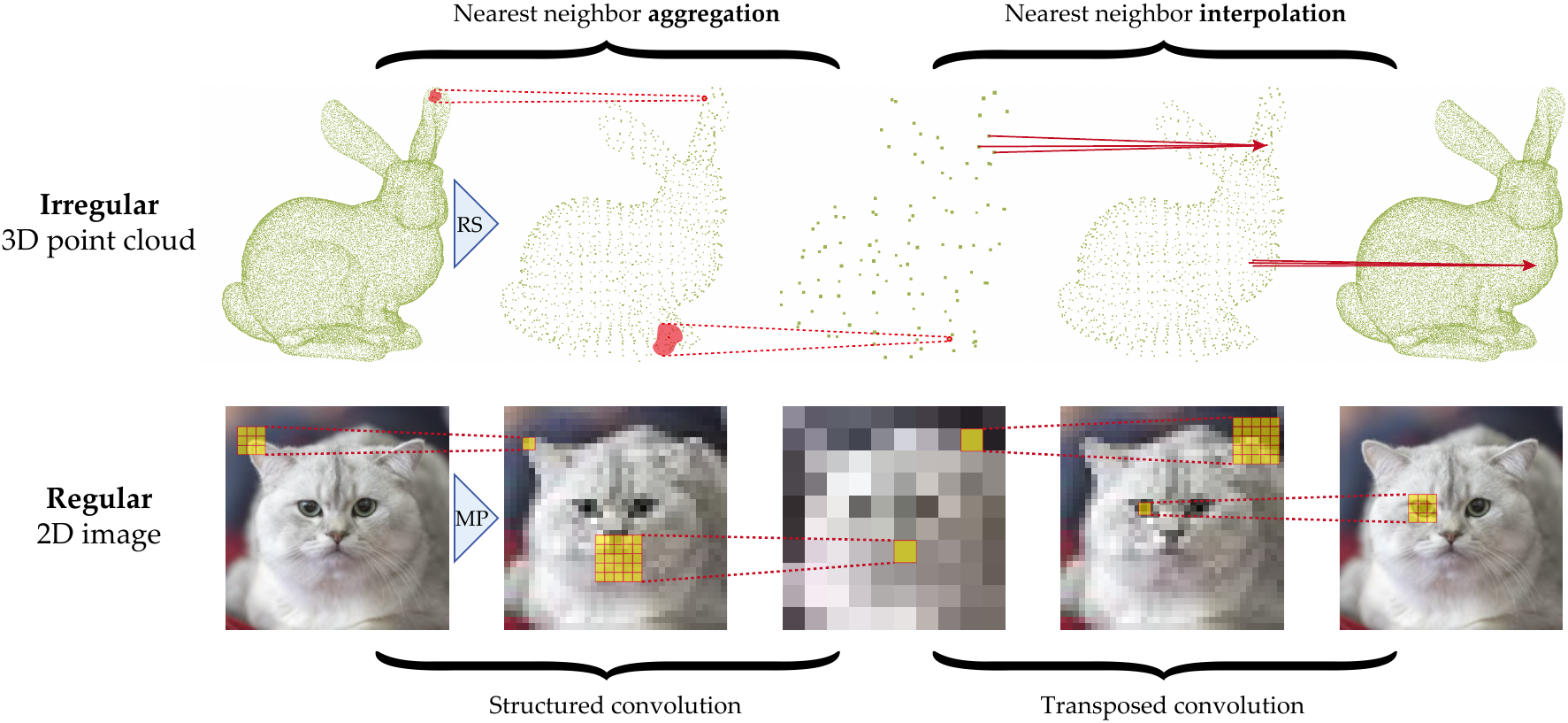}
\caption{Illustration of irregular 3D point cloud processing and regular 2D image processing. RS: random sampling; MP: max pooling. Typically, max pooling is used to down-sample regular 2D images, in this process, structured CNNs are used to capture local feature maps; random sampling or farthest point sampling is used to down-sample irregular 3D point clouds, in this process, unstructured nearest neighbor aggregation are used to capture local feature maps. Then, structured transposed CNNs and pixel shuffle are usually used for 2D images to recover high-resolution information, whereas uneven nearest neighbor interpolations are usually used for 3D point clouds. }
\label{fig:2D_3D}
\end{figure*}

\noindent \textbf{Comparison of network architecture between 2D and 3D. } As shown in Figure~\ref{fig:2D_3D}, we further provide an intuitive comparison of architecture and operation between 2D and 3D networks. Clearly, the input 2D images are structured and regular, while 3D point clouds are discrete and irregular. As such, although aiming for the same segmentation task, the following differences exist: \textbf{(1) Sampling}. Max-pooling with strides is usually used to reduce the resolution in 2D images, while random or farthest point sampling are usually used to down-sample the 3D point clouds; \textbf{(2) Neighborhood Definition}. Considering the non-uniform distribution and uneven density of 3D point clouds, KNN and spherical search are usually used to determine the neighborhood, while the neighborhood can be easily determined by the neighborhood pixels in 2D images;  \textbf{(3) Low-level operations}. Standard CNN is used as the fundamental operation for 2D segmentation, while shared MLPs are usually taken as the basic operation in 3D point clouds. Overall, considering the orderless and unstructured nature of 3D point clouds, coupled with unstable neighborhoods and aggressive down-sampling operations, it should be more careful to aggregate the multi-scale features from different nodes in U-Net architecture. Motivated by this, we go back to the beginning and stack as many U-Net $L^1$ codecs as possible in our U-Next framework, and introduce multi-level deep supervision, to ensure that the aggregated features have minimal semantic gaps, further leading to increasing performance in semantic segmentation task.

\begin{figure}[]
\setlength{\abovecaptionskip}{0.1cm}
\centering
\includegraphics[width=0.45\textwidth]{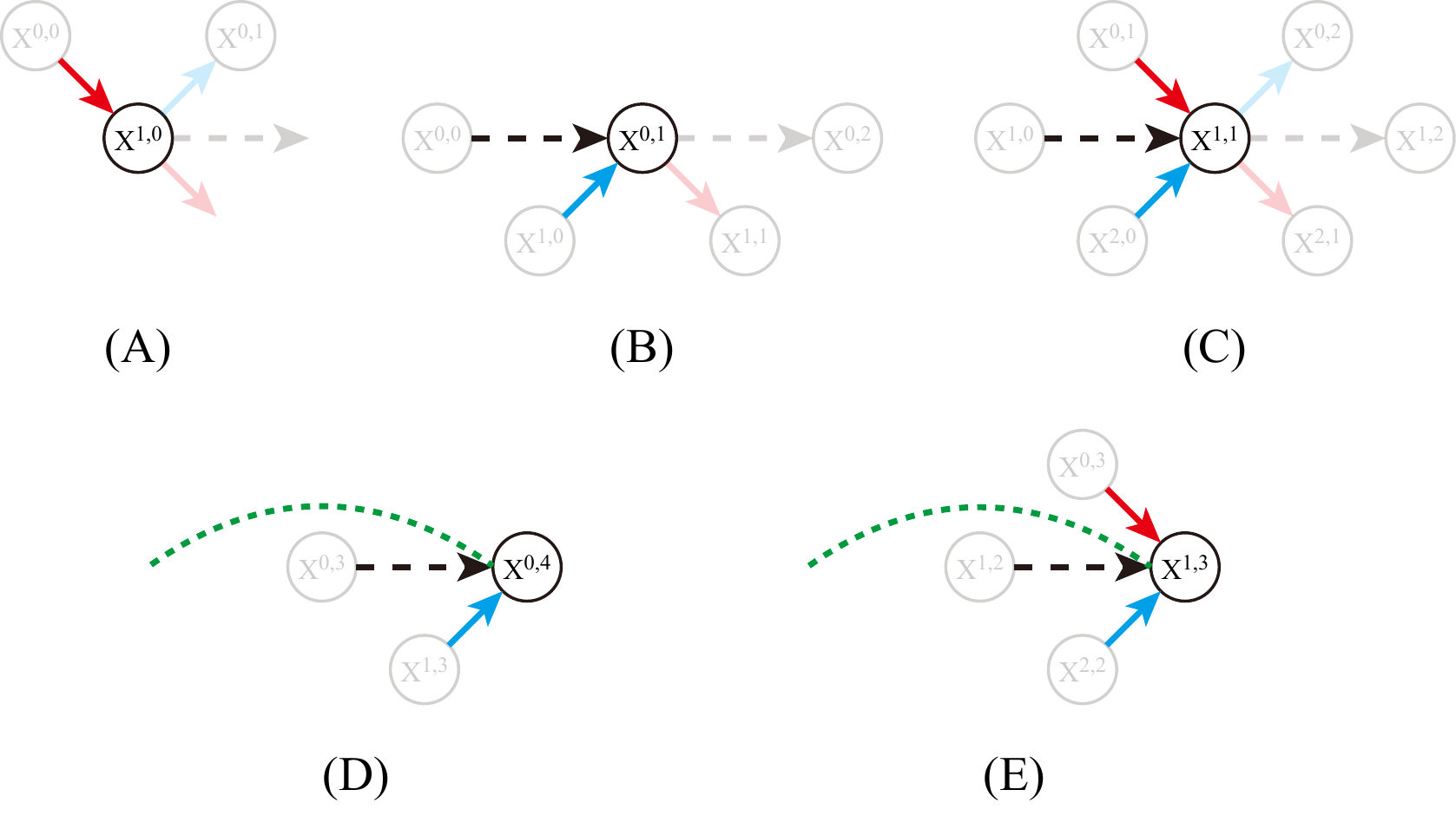}
\caption{Node connectivity pipeline for different types of nodes in U-Net$+^d$ and U-Next. The input of each node type is prominently identified. U-Net$+^d$ is only represented in (A-C), and U-Next builds on the U-Net$+^d$ also includes (D-E).}
\label{fig:input_type}
\end{figure}

\vspace{1.5mm}
\noindent \textbf{Node Connectivity Pipeline.} To further illustrate how we aggregate features from different nodes, we use pseudo-code to show the node connectivity pipeline in the U-Next architecture, as shown in Algorithm~\ref{algo: Node Connectivity Pipeline}. Basically, as shown in Figure~\ref{fig:input_type} (A-C), the node with $j=0$ receives only the output of the previous coding layer as input, the node with $i=0\ (j>0)$ receives two inputs, both from the same U-Net $L^1$ sub-network, and the node with $i>0$ and $j>0$ receives the output of the previous coding layer as input, and also the two inputs from the same U-Net $L^1$ sub-network. U-Next introduces additional long skip connections on top of U-Net+$^d$, instead of the dense skip connections of U-Net++, considering that there is no apparent performance improvement (validated in Section~\ref{sec:abla}). Figure~\ref{fig:input_type} (D-E) further clarifies the stack of feature maps in additional long skip connections represented by $x^{i,j}$, if and only if node $X^{i,j}$ is the last node in $i$-th row.

\IncMargin{1em}
\begin{algorithm} \SetKwInOut{Input}{input}\SetKwInOut{Output}{output}
	\Input{Initial embedding} 
	\Output{Recovered ori-resolution feature map}
	 \BlankLine 
	 \For{$j$ in range($N$)}{ 
	 \label{algo: Node Connectivity Pipeline}
	 	\If {j==0}{\label{lt} 
 		 	\For{$i$ in range($N-j$)}{\label{forins}
	 			x = \textbf{Coding}(x)\;
	 			\textbf{List }[$j$] $\leftarrow$ x\;
	 			x = DownSampling(x)\;
	 			}
	 	}
 		 \lElse{
     		 \If {j$>$0}{
  		 	 	\For{$i$ in range($N-j$)}{\label{forins}
         		 x0 = \textbf{List }[$j-1$][$i$]\;
         		 x1 = UpSampling(\textbf{List }[$j-1$][$i+1$])\;
         		 \If {i$>$0}{
             		 x2 = DownSampling(x)\;
     		    }
         		 \If {i+j==N-\rm{1}}{
         		 \textit{// The last Node of each layer}\\
             		 x3 = \textbf{List }[$0$][$i$]\;
     		    }
      		    x = concatenate all the above\;
     			x = \textbf{Coding}(x)\;
     			\textbf{List }[$j$] $\leftarrow$ x\;
 		    }
 		}
 	 	\textit{// switch to next Sequence}
 		}
 	 } 
  	 \caption{Node Connectivity Pipeline}
\end{algorithm}
\DecMargin{-2em}

\begin{figure}[]
\centering
\setlength{\abovecaptionskip}{0.1cm}
\includegraphics[width=0.45\textwidth]{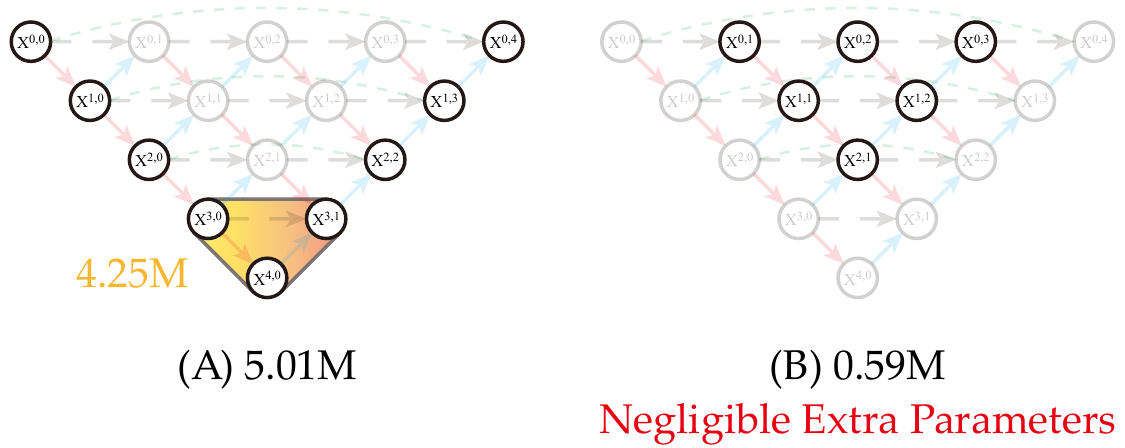}
\caption{Illustration of parameters (millions) in U-Next architecture. (A) shows parameters of "U-Net" backbone in U-Next. (B) shows parameters of additional nodes compared to "U-Net" backbone in U-Next.}
\label{fig:param}
\vspace{-1.0em} 
\end{figure}

\begin{table}[]
\setlength{\abovecaptionskip}{0cm}
\renewcommand\arraystretch{1.35}
\centering
\caption{The output of points and feature dimensions of coding blocks in our U-Next. The output size of each node in the same layer is the same.}
\setlength{\tabcolsep}{5mm}{
\begin{tabular}{ccc}
\bottomrule
Layer & Points & Dimension \\
\hline
$L_0$ & $\frac{N}{4}$ & 16\\
$L_1$ & $\frac{N}{16}$ & 64\\
$L_2$ & $\frac{N}{64}$ & 128\\
$L_3$ & $\frac{N}{256}$ & 256 \\
$L_4$ & $\frac{N}{512}$ & 512\\
\bottomrule
\end{tabular}
\label{tab:output}}%
\vspace{-1.0em}
\end{table}

\section{Experimental Results}
\label{sec:exp_re}

\subsection{Evaluation Metrics}
To quantitatively analyze the performance of the proposed architecture, overall accuracy (OA), per-class intersection over union (IoUs), and mean IoU (mIoU),  are used as evaluation metrics as follows:

{\setlength\abovedisplayskip{0pt}
\begin{align}
\mathrm{OA=\frac{\sum_{i=1}^{n}TP_i}{N}}
\end{align}
}

{\setlength\abovedisplayskip{0pt}
\begin{align}
\mathrm{IoU_i=\frac{TP_i}{TP_i+FP_i+FN_i}}
\end{align}
}

{\setlength\abovedisplayskip{0pt}
\begin{align}
\mathrm{mIoU=\frac{\sum_{i=1}^{n}{IoU}_i}{n}}
\end{align}
}

\noindent where $\mathrm{TP}$ represents the number of true positive samples, $\mathrm{FP}$ represents the number of false positive samples, $\mathrm{FN}$ represents the number of false negative samples, $i$ represents the ${i}_{th}$ semantic class, ${n}$ represents the number of semantic classes and $\mathrm{N}$ represents the number of total points.

\subsection{Additional Quantitative Results}
\noindent \textbf{Quantitative experiments on S3DIS Area5. }We also report the quantitative results of our U-Nextf framework with RandLA-Net \cite{RandLA-Net} and LACV-Net \cite{LACV-Net} as baselines on S3DIS Area 5 in Table~\ref{tab:s3dis_5}. From the result, we can see that by replacing the original U-Net \cite{UNet} with our U-Next, the overall segmentation performance can be significantly improved. We also noticed that our U-Next with LACV-Net almost achieves the top performance, except for slightly inferior to PointNeXt-XL \cite{PointNext}, which uses many training tricks including data augmentation and farthest point sampling with heavy pre/post-processing steps. It is also noted that PointNeXt-XL has 41.6M parameters, which is more than seven times than our U-Next (LACV-Net).

\begin{table}[]
\setlength{\abovecaptionskip}{0cm}
\setlength{\belowcaptionskip}{-0.2cm}
\centering
\vspace{0.5em}
\caption{Quantitative comparisons with the state-of-the-art methods on S3DIS (\textit{Area 5}).}
\scalebox{0.8}{
    \begin{tabular}{c|rccc}
    \bottomrule
    \multirow{2}*{\textbf{year}} & \multirow{2}*{\textbf{Method}} & \multirow{2}*{\textbf{OA}} & \multirow{2}*{\textbf{mAcc}} & \multirow{2}*{\textbf{mIoU}} \\
    &&&&\\
    \hline
    2017 & PointNet \cite{PointNet}      &  -   & 49.0 & 41.1\\
    2018 & DGCNN \cite{DGCNN}            & 83.6 &  -   & 47.9\\
    2019 & KPConv \cite{KPConv}          &  -   & 72.8 & 67.1\\    
    2021 & BAAF-Net \cite{BAAF-Net}      & 88.9 & 73.1 & 65.4\\
    2021 & PCT \cite{PCT}                &  -   & 67.7 & 61.3\\
    2021 & Point Transformer \cite{PT}   & 90.8 & 76.5 & 70.4\\    
    2022 & CBL \cite{CBL}                & 90.6 & 75.2 & 69.4\\
    2022 & PViT \cite{PViT}              &  -   & 75.2 & 69.6\\
    2022 & PointNeXt-L \cite{PointNext}  & 90.0 & 75.3 & 69.0\\
    2022 & PointNeXt-XL \cite{PointNext} & 90.6 & 76.8 & 70.5\\
    \hline
    2020 & RandLA-Net \cite{RandLA-Net}  & 87.2 & 71.4 & 62.5\\
    Ours & RandLA-Net \textcolor{red}{+U-Next} & 89.5 \textcolor{red}{+2.3} & 76.2 \textcolor{red}{+4.8} & 68.3 \textcolor{red}{+5.8}\\
    \hline
    2022 & LACV-Net \cite{LACV-Net}      & 89.3 & 73.6 & 65.9\\
    Ours & LACV-Net \textcolor{red}{+U-Next}   & 90.1 \textcolor{red}{+1.8} & 77.1 \textcolor{red}{+3.5} & 69.9 \textcolor{red}{+4.0} \\
    \bottomrule
    \end{tabular}
    \label{tab:s3dis_5}}
\vspace{-1.0em}
\end{table}

\begin{table*}[]
\setlength\tabcolsep{10pt}
\setlength{\abovecaptionskip}{0cm}
\setlength{\belowcaptionskip}{-0.2cm}
\centering
\caption{Detailed quantitative results of different approaches on Toronto3D dataset (\textit{Area 2}).}
\scalebox{0.8}{
\begin{tabular}{crcccccccccccccc}
\bottomrule
\multirow{2}*{RGB}&\multirow{2}*{Method}& \multirow{2}*{OA}& \multirow{2}*{mIoU}& \multirow{2}*{Road}& \multirow{2}*{Road mrk.}& \multirow{2}*{Nature}& \multirow{2}*{Buil.}& \multirow{2}*{Util. line}& \multirow{2}*{Pole}& \multirow{2}*{Car}& \multirow{2}*{Fence}\\
&&&&&&&&&&\\
\hline
\multirow{9}*{No}&PointNet++ \cite{PointNet++} & 92.6 & 59.5 & 92.9 & 0.0 & 86.1 & 82.2 & 60.9 & 62.8 & 76.4 & 14.4 \\
~&DGCNN \cite{DGCNN} & 94.2 & 61.7 & 93.9 & 0.0 & 91.3 & 80.4 & 62.4 & 62.3 & 88.3 & 15.8 \\
~&MS-PCNN \cite{MS-PCNN} & 90.0 & 65.9 & 93.8 & 3.8 & 93.5 & 82.6 & 67.8 & 71.9 & 91.1 & 22.5 \\
~&KPConv \cite{KPConv} & 95.4 & 69.1 & 94.6 & 0.1 & 96.1 & 91.5 & 87.7 & \textbf{81.6} & 85.7 & 15.7 \\
~&TGNet \cite{TGNet} & 94.1 & 61.3 & 93.5 & 0.0 & 90.8 & 81.6 & 65.3 & 62.9 & 88.7 & 7.9 \\
~&MS-TGNet \cite{Toronto3D} & 95.7 & 70.5 & 94.4 & 17.2 & 95.7 & 88.8 & 76.0 & 73.9 & 94.2 & 23.6 \\
~&LACV-Net \cite{LACV-Net} & 95.8 & 78.5 & 94.8 & 42.7 & 96.7 & 91.4 & \textbf{88.2} & 79.6 & 93.9 & \textbf{40.6}\\
\cline{2-12}
~&RandLA-Net \cite{RandLA-Net} & 93.0 & 77.7 & 94.6 & 42.6 & 96.9 & 93.0 & 86.5 & 78.1 & 92.9 & 37.1\\
~&\textcolor{red}{+ U-Next} & \textbf{96.0} \textcolor{red}{+3.0} & \textbf{79.2} \textcolor{red}{+1.5} & \textbf{95.1} & \textbf{44.8} & \textbf{97.2} & \textbf{93.5} & 87.6 & 80.5 & \textbf{94.3} & 40.8\\
\hline
\multirow{6}*{Yes}&ResDLPS-Net \cite{ResDLPS-Net} & 96.5 & 80.3 & 95.8 & 59.8 & 96.1 & 90.9 & 86.8 & 79.9 & 89.4 & 43.3\\
~&BAF-LAC \cite{BAF-LAC} & 95.2 & 82.2 & 96.6 & 64.7 & 96.4 & 92.8 & 86.1 & 83.9 & 93.7 & 43.5\\
~&BAAF-Net \cite{BAAF-Net} & 94.2 & 81.2 & 96.8 & 67.3 & 96.8 & 92.2 & 86.8 & 82.3 & 93.1 & 34.0\\
~&LACV-Net \cite{LACV-Net} & 97.4 & 82.7 & 97.1 & 66.9 & 97.3 & 93.0 & 87.3 & 83.4 & 93.4 & 43.1\\
\cline{2-12}
~&RandLA-Net \cite{RandLA-Net} & 94.4 & 81.8 & 96.7 & 64.2 & 96.9 & 94.2 & 88.0 & 77.8 & 93.4 & 42.9 \\
~&\textcolor{red}{+ U-Next} & \textbf{97.7} \textcolor{red}{+3.3} & \textbf{84.0} \textcolor{red}{+2.2} & \textbf{97.3} & \textbf{68.6} & \textbf{97.7} & \textbf{95.2} & \textbf{88.4} & \textbf{86.1} & \textbf{95.1} & \textbf{43.2}\\
\bottomrule
\end{tabular}%
\label{tab:toronto_de}}
\vspace{-1.0em}
\end{table*}

\vspace{1.5mm}
\noindent \textbf{Detailed quantitative results on Toronto3D. }
We report the detailed quantitative results achieved by our U-Next (RandLA-Net) on Toronto3D Area 2 in Table~\ref{tab:toronto_de}. It can be seen that our U-Next framework can steadily improve the segmentation performance of RandLA-net in all of the classes. In particular, the performance of our method is superior to the baseline in classes with vague local geometric information such as road markings and poles. This is primarily because our framework stacks more sub-networks to capture sufficient local details.

\vspace{1.5mm}
\noindent \textbf{Quantitative results on ScanNet v2.}
The ScanNet v2 dataset comprises 1,513 indoor room scans reconstructed from RGB-D frames. The dataset is partitioned into 1,201 scenes for training and 312 for validation, and all are annotated with 20 semantic classes. In this study, we evaluate the performance of our proposed U-Next (RandLA-Net) on the ScanNet v2 validation set and report the quantitative results in Table~\ref{tab:scannet}.
The experimental results demonstrate that by replacing the original U-Net \cite{UNet} with our U-Next, the overall segmentation performance of both baseline networks can be significantly improved. Specifically, the proposed method achieves an improvement of 6.2\% and 5.7\% in mIoU scores compared to the baseline RandLA-Net and LACV-Net, respectively. Notably, the U-Next (LACV-Net) achieves state-of-the-art (SOTA) performance on the validation set of the ScanNet v2 dataset, demonstrating the effectiveness of our proposed U-Next framework.

\vspace{1.5mm}
\noindent \textbf{Quantitative experiments on Instance Segmentation. }
To investigate the versatility and generalizability of our proposed U-Next framework beyond the semantic segmentation task, we conduct further experiments on the task of instance segmentation. Specifically, we re-implement 3D-BoNet \cite{3D-BoNet} using our U-Next architecture and evaluate its performance on Area 5 of the S3DIS dataset. We report the quantitative results using mean class precision (mPrec) and mean class recall (mRec) as evaluation metrics in Table~\ref{tab:ins}. The results demonstrate that our U-Next architecture can significantly improve instance segmentation performance in addition to semantic segmentation. Specifically, the proposed method achieves an improvement of 5.6\% and 9.4\% in mPrec and mRec scores, respectively, compared to the baseline 3D-BoNet. This finding indicates that our U-Next framework can generalize well across different tasks and highlights its potential for various computer vision tasks.

\vspace{1.5mm}
\noindent \textbf{Trainable parameters. }
In Table~\ref{tab:arc} of the main paper, we report the trainable parameters of the U-Next. Here we further provide detailed parameters in different layers of the architecture, as shown in Figure~\ref{fig:param}. It can be seen that the deepest U-Net $L^1$ sub-network occupies the largest number of parameters compared to the other parts (about 75.76\% of the total parameters), and the parameters of additional nodes in U-Next compared to the ”U-Net” backbone only occupy a small part of parameters (about 10.57\%). Therefore, the proposed U-Next does not introduce too many extra parameters compared to the original U-Net.

\begin{table}[]
\setlength{\abovecaptionskip}{0cm}
\begin{center}
\caption{Quantitative comparisons with the state-of-the-art methods on ScanNetv2.}
\scalebox{0.8}{
    \begin{tabular}{rc}
    \bottomrule
    \multirow{2}*{\textbf{Method}} & \textbf{val mIoU}\\
    &(\%)\\
    \hline
    PointNet++ \cite{PointNet++} & 53.5\\
    PointConv \cite{PointConv} & 61.0\\
    PointASNL \cite{PointASNL} & 63.5\\
    KPConv \cite{KPConv} & 69.2\\
    SparseConvNet \cite{SparseConv} & 69.3\\
    PointTransformer \cite{PT} & 70.6\\
    \hline
    RandLA-Net \cite{RandLA-Net} & 63.5\\
    RandLA-Net \textcolor{red}{+U-Next} & 69.7 \textcolor{red}{+6.2}\\
    \hline
    LACV-Net \cite{LACV-Net} & 66.1\\
    LACV-Net \textcolor{red}{+U-Next} & 71.8 \textcolor{red}{+5.7}\\
    \bottomrule
    \end{tabular}}
    \label{tab:scannet}
\end{center}
\vspace{-2.0em}
\end{table}

\begin{table}[]
\setlength{\abovecaptionskip}{0cm}
\begin{center}
\caption{Quantitative comparisons of \textbf{instance segmentation} results ($\%$) on S3DIS (\textit{Area 5}). * is our implementation.}
\scalebox{0.8}{
    \begin{tabular}{rccc}
    \bottomrule
    \multirow{2}*{\textbf{Architecture}} & \textbf{mPrec} & \textbf{mRec} \\
    &(\%)&(\%)\\
    \hline
    ASIS \cite{ASIS} & 55.3 & 42.4\\
    3D-BoNet \cite{3D-BoNet}  & 57.5 & 40.2\\
    PN++ \cite{PN++} & 60.1 & 47.2\\
    InsEmb \cite{InsEmb} & 61.3 & 48.5\\
    \hline
    3D-BoNet*  & 55.2 & 39.1\\
    3D-BoNet \textcolor{red}{+ U-Next} & 60.8 \textcolor{red}{+5.6} & 48.5 \textcolor{red}{+9.4}\\
    \bottomrule
    \end{tabular}}
    \label{tab:ins}
\end{center}
\vspace{-1.0em}
\end{table}

\subsection{Additional ablation studies}
\label{sec:abla}
In this section, we conduct ablation studies on the proposed U-Next in the following four aspects. All the ablation networks are trained on Areas 1-4 and 6, and tested on Area 5 of the S3DIS \cite{s3dis} dataset. RandLA-Net \cite{RandLA-Net} is used as the main baseline model.

\vspace{1.5mm}
\noindent \textbf{Ablation study of different levels of U-Next. }
Here, we compare the performance and number of parameters of U-Next with different levels. Reducing the number of layers of U-Next can significantly reduce the inference time, but segmentation performance also degrades. As such, the level of U-Next should be a trade-off between computational costs and accuracy. As shown in Table~\ref{tab:UNext_level}, U-Next $L^3$ reduces the memory footprint (number of parameters) by 74.6\%, while only reducing the mIoU score by 2.1\%. More aggressive levels further reduce memory footprint, but at the cost of significant accuracy degradation. On the other hand, we also conducted similar experiments on U-Net with different levels, as shown in Table~\ref{tab:UNet_level}. It can be seen that the higher the level, the greater the performance improvement of our U-Next compared with the U-Net framework, primarily because more sub-networks are stacked as the level increase, which also means the more sufficient feature fusion.

\begin{table}[]
\setlength{\abovecaptionskip}{0cm}
\setlength{\belowcaptionskip}{-0.2cm}
\begin{center}
\caption{Semantic segmentation results ($\%$) and parameters (millions) of different level of U-Next on S3DIS (\textit{Area 5}).}
\scalebox{0.8}{
    \begin{tabular}{rccc}
    \bottomrule
    \multirow{2}*{\textbf{Architecture}} & \textbf{OA} & \textbf{mIoU} & \textbf{Parameters}\\
    &(\%)&(\%)&(millions)\\
    \hline
    U-Next $L^4$    & 89.5 & 68.3 & 5.61 \\
    U-Next $L^3$    & 89.1 & 66.1 & 1.43 \\
    U-Next $L^2$    & 86.8 & 60.6 & 0.31 \\
    U-Next $L^1$    & 81.0 & 49.6 & 0.05 \\
    \bottomrule
    \end{tabular}}
    \label{tab:UNext_level}
\end{center}
\vspace{-2.0em}
\end{table}

\begin{table}[]
\setlength{\abovecaptionskip}{0cm}
\setlength{\belowcaptionskip}{-0.2cm}
\begin{center}
\caption{Semantic segmentation results ($\%$) and parameters (millions) of different level of U-Net on S3DIS (\textit{Area 5}).}
\scalebox{0.8}{
    \begin{tabular}{rccc}
    \bottomrule
    \multirow{2}*{\textbf{Architecture}} & \textbf{OA} & \textbf{mIoU} & \textbf{Parameters}\\
    &(\%)&(\%)&(millions)\\
    \hline
    U-Net $L^4$    & 87.2 & 62.5 & 4.99 \\
    U-Net $L^3$    & 86.1 & 61.2 & 1.24 \\
    U-Net $L^2$    & 84.7 & 58.8 & 0.30 \\
    U-Net $L^1$    & 80.7 & 49.1 & 0.05 \\
    \bottomrule
    \end{tabular}}
    \label{tab:UNet_level}
\end{center}
\vspace{-1.0em}
\end{table}

\begin{table}[]
\setlength{\abovecaptionskip}{0cm}
\setlength{\belowcaptionskip}{-0.2cm}
\begin{center}
\caption{Semantic segmentation results ($\%$) and parameters (millions) of different architectures on S3DIS (\textit{Area 5}).}
\scalebox{0.8}{
    \begin{tabular}{rccc}
    \bottomrule
    \multirow{2}*{\textbf{Architecture}} & \textbf{OA} & \textbf{mIoU} & \textbf{Parameters}\\
    &(\%)&(\%)&(millions)\\
    \hline
    wide U-Net    & 87.4 & 63.2 & 5.64 \\
    wide U-Net++  & 87.8 & 64.9 & 6.09 \\
    \textbf{U-Next}        & 89.5 & 68.3 & 5.61 \\
    \bottomrule
    \end{tabular}}
    \label{tab:wide}
\end{center}
\vspace{-1.0em}
\end{table}

\begin{figure*}[]
\setlength{\abovecaptionskip}{0.1cm}
\centering
\includegraphics[width=0.95\textwidth]{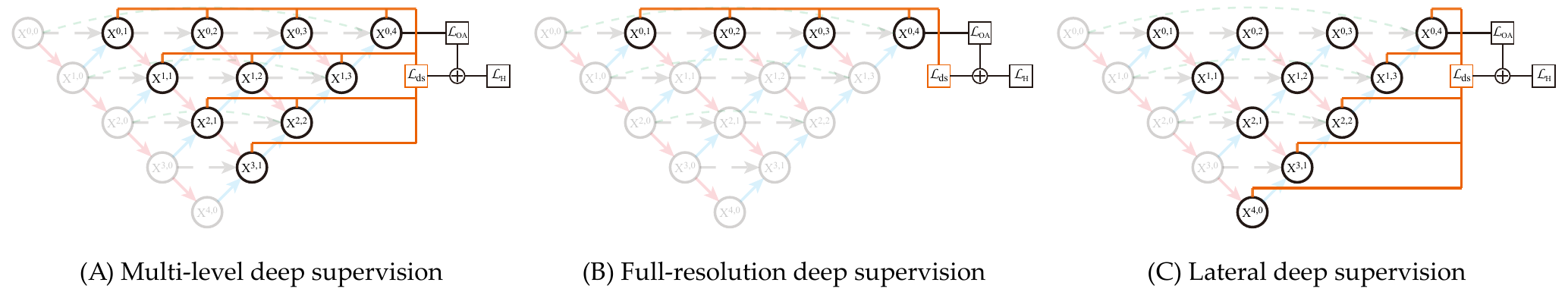}
\caption{Illustration of the proposed multi-level deep supervision, the full-resolution deep supervision proposed by \cite{UNet++}, and the lateral deep supervision proposed by \cite{UNet3}.}
\label{fig:dsc}
\vspace{-0.5em}
\end{figure*}

\vspace{1.5mm}
\noindent \textbf{Ablation study of wide U-Net and U-Net++. }
Compared with the previous U-Net \cite{UNet} and U-Net++ \cite{UNet++} framework, the proposed U-Next indeed introduces additional parameters. To verify that the performance improvement of U-Next is not attributed to additional parameters, but to the inherent advantages of the network architecture, we conducted the following experiments. We artificially increase the number of parameters of U-Net and U-Net++ by increasing the feature output dimension of the encoder and decoder. Specifically, the channel dimension of the features increases as: ($16 + 2\rightarrow 64 + 4\rightarrow 128 + 8\rightarrow 256 + 16\rightarrow 512 + 32$). Table~\ref{tab:wide} shows the results of the quantitative analysis, it is clear that although introducing additional parameters can improve the segmentation performance, the segmentation performance of the proposed U-Next is still significantly superior to wide U-Net and U-Net++, which verifies that the performance improvements of the proposed U-Next are mainly due to the advantages of the network architecture.

\begin{table}[]
\setlength{\abovecaptionskip}{0cm}
\setlength{\belowcaptionskip}{-0.2cm}
\begin{center}
\caption{Semantic segmentation results ($\%$) and parameters (millions) of U-Next with different skip connections on S3DIS (\textit{Area 5}). conn.: connections.}
\scalebox{0.8}{
    \begin{tabular}{rccc}
    \bottomrule
    \multirow{2}*{\textbf{Architecture}} & \textbf{OA} & \textbf{mIoU} & \textbf{Parameters}\\
    &(\%)&(\%)&(millions)\\
    \hline
    w/o skip conn.   & 89.2 & 67.4 & 5.51\\
    Dense skip conn. & 89.4 & 67.9 & 5.66\\
    \textbf{Long skip conn.}  & 89.5 & 68.3 & 5.61 \\
    \bottomrule
    \end{tabular}}
    \label{tab:skip}
\end{center}
\vspace{-2.0em}
\end{table}

\begin{table}[]
\setlength{\abovecaptionskip}{0cm}
\setlength{\belowcaptionskip}{-0.2cm}
\begin{center}
\caption{Semantic segmentation results ($\%$) of U-Next with different deep supervision on S3DIS (\textit{Area 5}). DS.:deep supervision.}
\scalebox{0.8}{
    \begin{tabular}{rccc}
    \bottomrule
    \multirow{2}*{\textbf{Architecture}} & \textbf{OA} & \textbf{mIoU} \\
    &(\%)&(\%)\\
    \hline
    w/o DS. & 89.0 & 66.7\\
    Full-resolution DS.  & 89.2 & 67.5\\
    Lateral DS.  & 89.2 & 67.7 &\\
    \textbf{Multi-level DS.} & 89.5 & 68.3\\
    \bottomrule
    \end{tabular}}
    \label{tab:ds}
\end{center}
\vspace{-2.0em}
\end{table}

\vspace{1.5mm}
\noindent \textbf{Ablation study of long skip connection. }The proposed U-Next uses the design of long skip connections. It is noted that U-Net++ utilizes dense skip connections to connect all nodes at the same level. To verify the effectiveness of our design of long skip connections (and why not choose the design of U-Net++), we performed this ablation study. In this experiment, dense long skip connections are integrated into U-Next. As shown in Table~\ref{tab:skip}, we can observe that although the network using dense skip connections have superior performance compared to networks without using long skip connections, it is comparable to the performance of using long skip connections. This may be because each sub-network can learn local features with different scales sufficiently, and the dense skip connections are not that necessary, but increase the risk of overfitting. In particular, dense skip connections introduce additional parameters and computational cost, resulting in redundant computations. Therefore, we finally adopt the vanilla long skip connections as used in the basic U-Net.

\vspace{1.5mm}
\noindent \textbf{Ablation study of multi-level deep supervision. }
The proposed U-Next introduces multi-level deep supervision from the perspective of optimization and learning in each decoder node. We note that U-Net++ \cite{UNet++} only applies deep supervision on the generated full-resolution feature maps, and U-Net3+ \cite{UNet3} only applies deep supervision on the  lateral nodes. To further verify the effectiveness of our multi-level deep supervision, we performed this ablation study (The comparison of the three is shown in Figure~\ref{fig:dsc}). In this experiment, full-resolution deep supervision used in U-Net++ and lateral deep supervision used in U-Net3+ are integrated into U-Next. Table~\ref{tab:ds} shows the results of the quantitative analysis, from the results we can observe that deep supervision in full-resolution and lateral have superior performance compared to the network without using deep supervision, but has inferior performance compared to deep supervision in multi-level. This is mainly because the proposed multi-level deep supervision trains each U-Net $L^1$ sub-network independently, but the full-resolution deep or lateral supervision only trains on the original resolution scale or the rightmost nodes, the supervised sub-networks are limited.

\section{Results Visualization}
\label{sec:vis}
In this section, we visually compare the semantic segmentation results achieved by baseline RandLA-Net \cite{RandLA-Net} and that with our U-Next on the used three large-scale benchmark datasets. In each visualization result, red boxes show the part of RandLA-Net where the segmentation is wrong or the boundary segmentation is not significant.

\begin{figure*}[]
\centering
\includegraphics[width=1.0\textwidth]{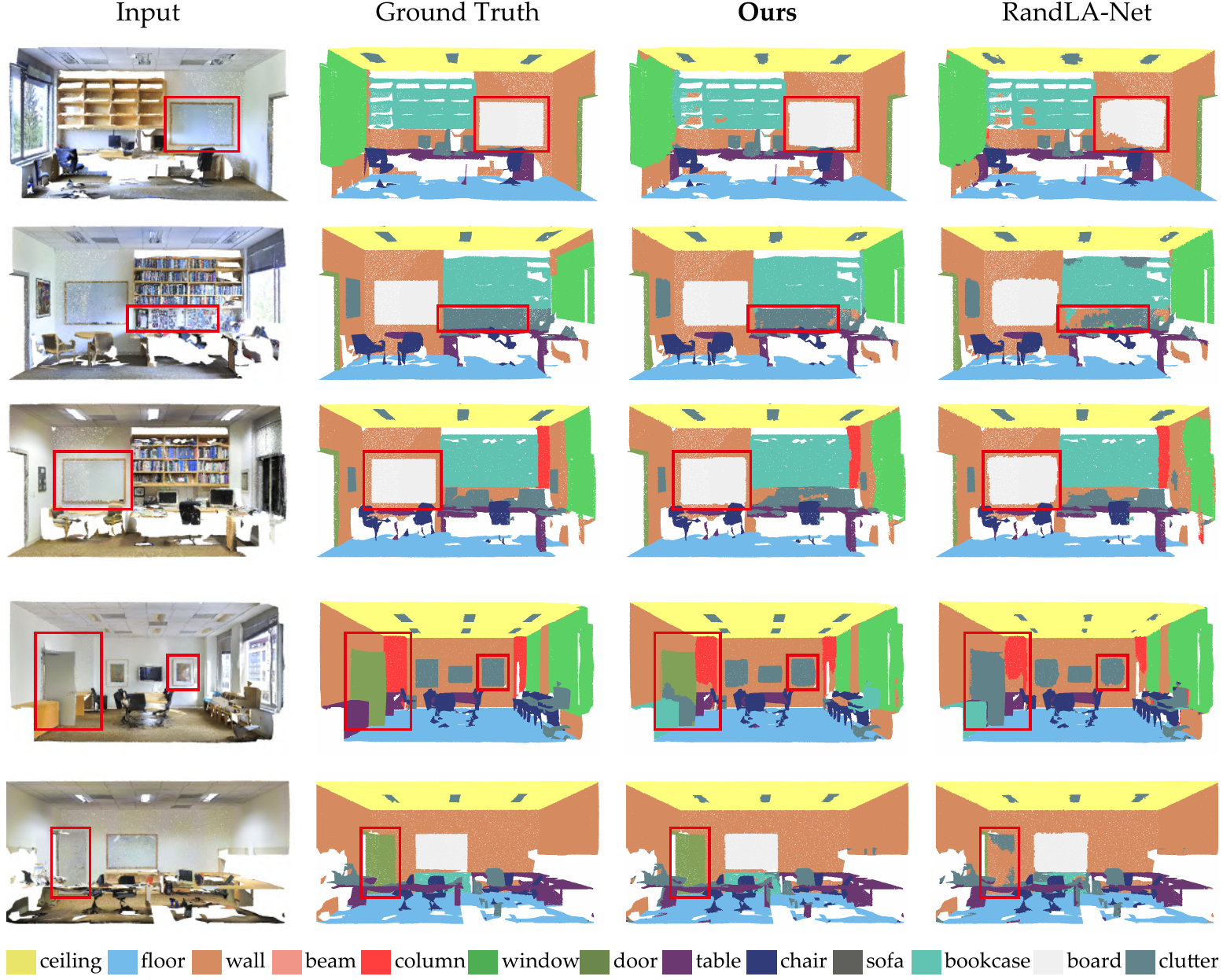}
\caption{Visual comparison of semantic segmentation results on S3DIS dataset.}
\label{fig:s3dis}
\end{figure*}

\vspace{1.5mm}
\noindent \textbf{S3DIS: }Figure~\ref{fig:s3dis} qualitatively shows the visual comparison on the S3DIS \cite{s3dis} dataset. From the result, it is visually clear that our segmentation of \textit{boards}, \textit{columns}, \textit{doors}, and \textit{bookcases} is superior to RandLA-Net. In addition, the proposed U-Next can segment the boundaries of objects more smoothly and accurately. This validates that repeated fusion of information from sub-networks with minimal semantic gaps helps to rebuild fine-grained perception to alleviate irregular and non-uniform sampling and aggregation in 3D point clouds. 

\begin{figure*}
\centering
\includegraphics[width=1.0\textwidth]{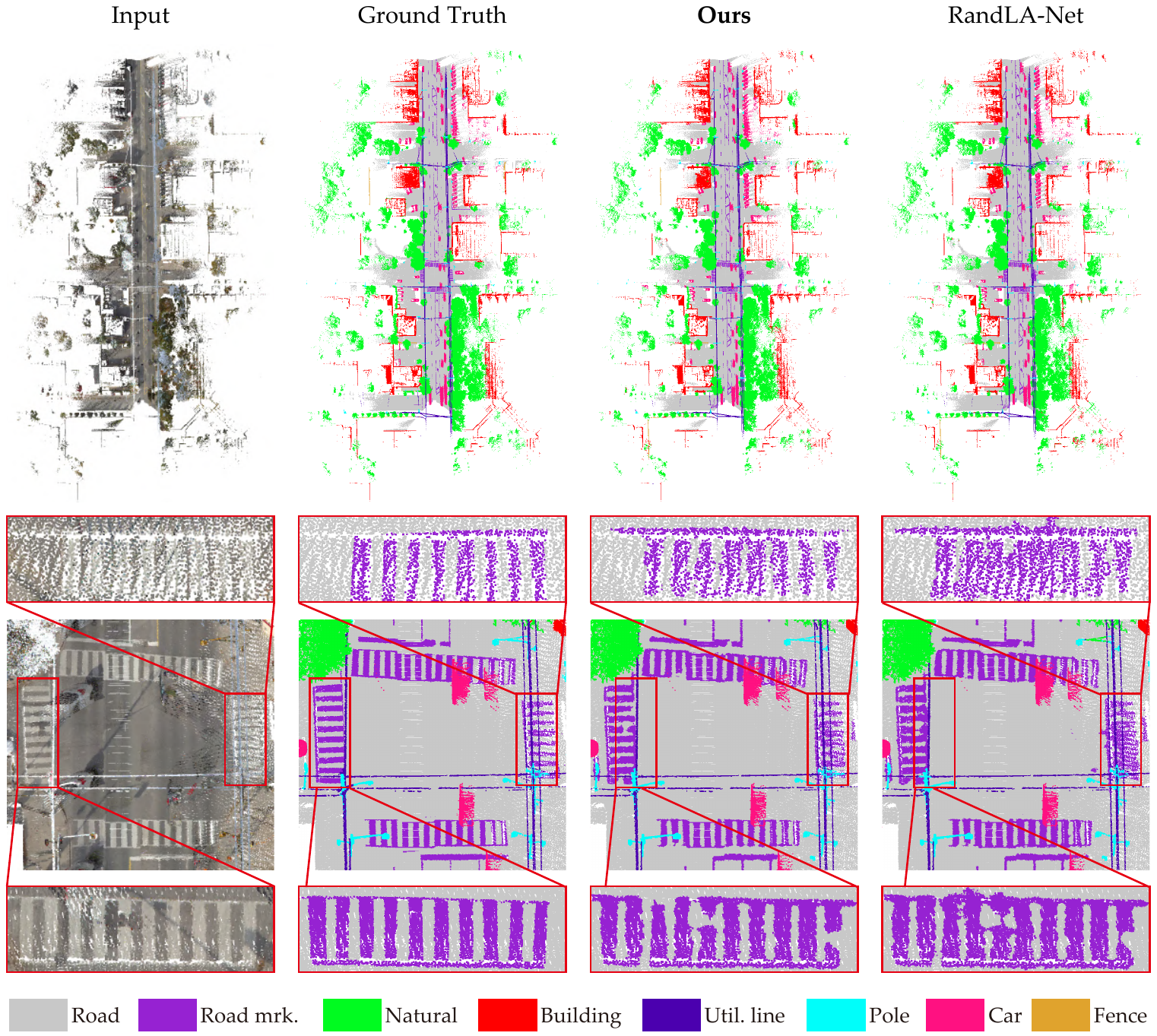}
\caption{Visual comparison of semantic segmentation results on Toronto3D dataset.}
\label{fig:toronto3D}
\end{figure*}

\vspace{1.5mm}
\noindent \textbf{Toronto3D: }Figure~\ref{fig:toronto3D} qualitatively shows the visual comparison on the Toronto3D \cite{Toronto3D} dataset using $rgb$ as initial input. From the results, we can observe that our segmentation of \textit{car}, \textit{natural} and \textit{fence} is significantly superior to that of RandLA-Net. In particular, for the semantic segmentation of the \textit{crosswalk}, the proposed U-Next is more accurate and clear in the boundaries, compared to the excellent RandLA-Net. 

\begin{figure*}
\centering
\includegraphics[width=1.0\textwidth]{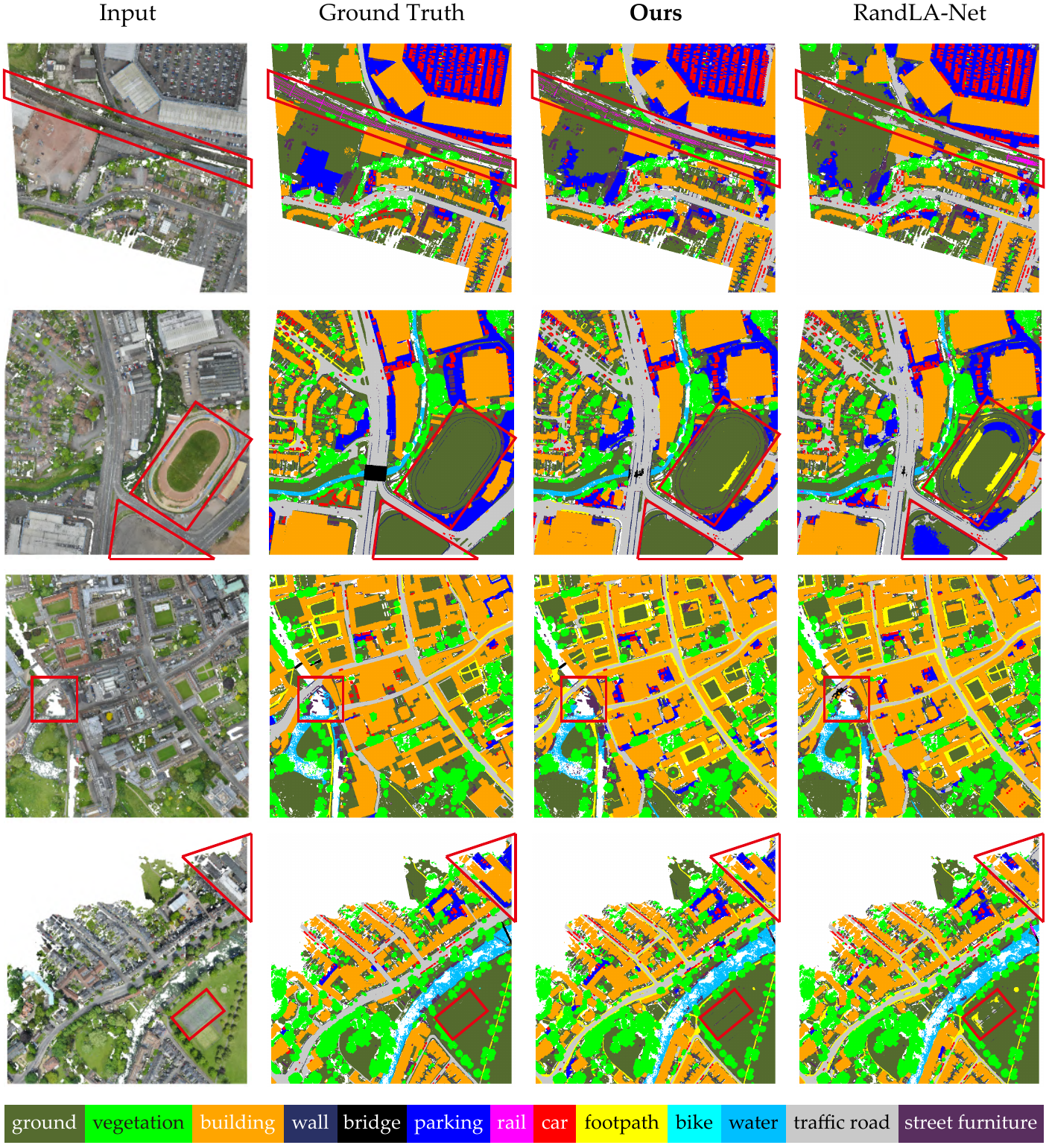}
\caption{Visual comparison of semantic segmentation results on SensatUrban dataset.}
\label{fig:SensatUrban}
\end{figure*}

\vspace{1.5mm}
\noindent \textbf{SensatUrban: }Figure~\ref{fig:SensatUrban} qualitatively shows the visual comparison on the SensatUrban \cite{SensatUrban} dataset. Since the ground truth of the online test sets is not publicly available, to better demonstrate the visualization, we show the results on the validation set. It can be observed that compared to RandLA-Net, which barely segments the \textit{railway}, the proposed U-Next can clearly segment the \textit{railway} from the ground. Meanwhile, it has a significant improvement in the performance of large \textit{parking} and small \textit{traffic road}. 

\clearpage

\clearpage

\clearpage

\end{document}